\documentclass[11pt]{article}
\usepackage[preprint]{acl}
 
\usepackage{times}
\usepackage{latexsym}
\usepackage[T1]{fontenc}
\usepackage[utf8]{inputenc}
\usepackage{microtype}
 
\usepackage{graphicx}
\usepackage{amsmath}
\usepackage{amssymb}
\usepackage{booktabs}
\usepackage{multirow}
\usepackage{enumitem}
\usepackage{xspace}
\usepackage{adjustbox}
\usepackage{makecell}
\usepackage{xcolor}
\usepackage{tcolorbox}
\usepackage{tikz}
\usetikzlibrary{positioning,shapes.geometric,shapes.misc,arrows.meta,fit,calc,backgrounds,decorations.markings}
\usepackage{colortbl}
\usepackage{array}
 
\definecolor{axAssetT}{HTML}{C0392B}
\definecolor{axOrchT}{HTML}{2E86AB}
\definecolor{axKnowT}{HTML}{1E8449}
\definecolor{axInfraT}{HTML}{D68910}
\definecolor{axReasT}{HTML}{7D3C98}
\definecolor{axEvalT}{HTML}{B9591E}
 
\newcommand{\axispill}[1]{%
  \tikz[baseline=-0.55ex]{\fill[#1] (0,0) rectangle (0.32cm,0.32cm);}%
}
 
\definecolor{tblHdrBg}{HTML}{1F4E79}
\definecolor{tblHdrFg}{HTML}{FFFFFF}
\definecolor{tblRowAlt}{HTML}{F2F4F8}
 
\newtcolorbox{position}{
  colback=blue!4,colframe=blue!50!black,
  boxrule=0.4pt,arc=2pt,
  left=4pt,right=4pt,top=3pt,bottom=3pt
}
\newtcolorbox{claim}{
  colback=red!4,colframe=red!50!black,
  boxrule=0.4pt,arc=2pt,
  left=4pt,right=4pt,top=3pt,bottom=3pt
}
 
\title{Beyond Static Leaderboards: \\ Predictive Validity for the Evaluation of LLM Agents}
 
\author{%
\large\bfseries Dhaval C.\ Patel\thanks{Corresponding author: \texttt{pateldha@us.ibm.com}}\\[2pt]
\normalfont
Kaoutar El Maghraoui \quad Shuxin Lin \quad Yusheng Li \quad Tianjun Feng \quad Chun-Yi Tsai \quad Yihan Sun \\
Wei Alexander Xin \quad Akshat Bhandari \quad Tanisha Rathod \quad Aaron Fan \quad Sanskruti Vijay Shejwal \\
Tomas Pasiecznik \quad Sagar Chethan Kumar \quad Tanmay Agarwal \quad Rohith Kanathur \quad Sam Colman \\
Amaan Sheikh \quad Dev Bahl \quad Ann Li \quad Krish Veera \quad Alimurtaza Mustafa Merchant \\
Shambhawi Baswaraj Bhure \quad Sajal Kumar Goyla \quad Chengrui Li \quad Kirthana Natarajan \quad Rui Li \\
Thomas Ajai \quad Rujing Li \quad Vivek G.\ Iyer \quad Sanjaii Vijayakumar \quad Yitong Bai \quad Ayal Yakobe \\
Darief Maes \quad Yassine Jebbouri \quad Tianyang Xu \quad Thai Quoc On \quad Vera Mazeeva \quad Winston Li \\
Yuval Shemla \quad Yeshitha Bhuvanesh \quad Rushin Bhatt \quad Siddharth Chethan Gowda \quad Alisha Vinod \\
Caroline Cahill \quad Shriya Aishani Rachakonda \quad Yunfeng Chen \quad Aryaman Agrawal \quad Aman Upganlawar \\
Mao Le Jonathan Ang \quad Yubin Sally Go \quad Madhav Rajkondawar \quad Yang-Jung Chen \quad Trisha Maturi \\
Ananya Kapoor \quad Andrew Li \quad Shrey Arora \quad Mana Abbaszadeh \quad Shen Li \quad Charles Xu \quad Byeolah Kwon%
}
 
\begin{document}
\maketitle
\begin{abstract}
Agent benchmarks are growing fast, but no single benchmark touches
more than four or five of the dimensions that deployment exposes.
This paper aggregates the largest coordinated deep-dive of one
MCP-based industrial-agent benchmark to date: fourteen parallel
implementation studies covering new asset classes (including a
multi-modal visual extension), alternative orchestrations, retrieval
strategies, reasoning modes, infrastructure optimizations, and
evaluation-methodology probes. Consolidating those
studies with seven prior agent benchmarks, we argue that
aggregate-score leaderboards systematically underspecify deployed-agent
evaluation. Rankings derived from aggregate scores do not transfer to
out-of-distribution settings; recent public-to-hidden competition
retrospectives provide direct empirical evidence of this rank
instability. We propose ranking configurations by predictive validity,
the correlation between in-sample and out-of-sample rank, rather than
in-sample mean, and report a twelve-tier measurement apparatus that
exposes the deployment-relevant dimensions HELM and its agent-era
successors collapse. The position is operationalized through three
falsifiable out-of-distribution criteria with explicit thresholds;
existing evidence partly supports it but is too thin to confirm. We
close with a pre-registered pilot design and a field-level vision for
what the next generation of agentic benchmarks should report.
\end{abstract}

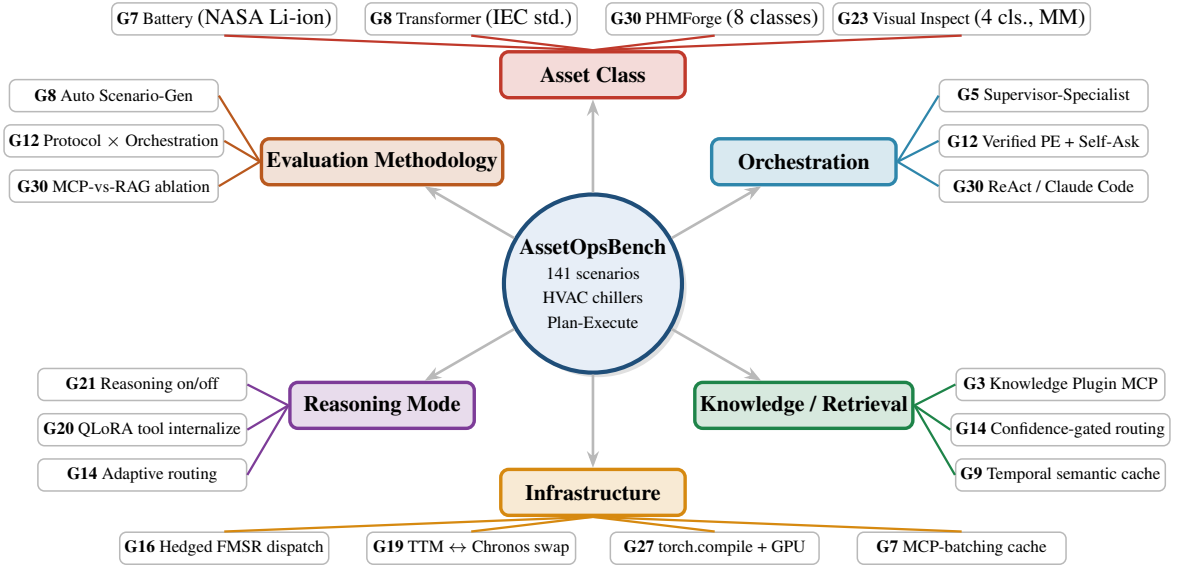
\begin{figure*}[t]
\centering
\definecolor{hubblue}{HTML}{1F4E79}
\definecolor{hubfill}{HTML}{E7EEF7}
\definecolor{axAsset}{HTML}{C0392B}
\definecolor{axOrch}{HTML}{2E86AB}
\definecolor{axKnow}{HTML}{1E8449}
\definecolor{axInfra}{HTML}{D68910}
\definecolor{axReas}{HTML}{7D3C98}
\definecolor{axEval}{HTML}{B9591E}
\begin{tikzpicture}[
  scale=0.92, transform shape,
  >={Stealth[length=2.2mm,width=1.6mm]},
  hubshadow/.style={
    circle, fill=black!12, minimum size=2.55cm, inner sep=1pt
  },
  hub/.style={
    circle, draw=hubblue, fill=hubfill, line width=0.55mm,
    minimum size=2.55cm, align=center, font=\small, inner sep=1pt
  },
  axislab/.style={
    rectangle, rounded corners=3pt, line width=0.45mm,
    minimum width=2.65cm, minimum height=0.66cm, align=center, font=\bfseries\footnotesize, inner sep=2pt
  },
  leaf/.style={
    rectangle, rounded corners=2.5pt, draw=gray!55, fill=white, line width=0.2mm,
    minimum width=3.00cm, minimum height=0.46cm, font=\scriptsize, align=center, inner sep=2pt
  },
]
\node[hubshadow] at (0.07,-0.07) {};
\node[hub] (aob) at (0,0) {\textbf{AssetOpsBench}\\\scriptsize 141 scenarios\\\scriptsize HVAC chillers\\\scriptsize Plan-Execute};

\node[axislab, fill=axAsset!18,  draw=axAsset]  (lAss) at ( 90:3.0) {Asset Class};
\node[axislab, fill=axOrch!18,   draw=axOrch]   (lOrc) at ( 30:3.5) {Orchestration};
\node[axislab, fill=axKnow!18,   draw=axKnow]   (lKno) at (-30:3.5) {Knowledge / Retrieval};
\node[axislab, fill=axInfra!18,  draw=axInfra]  (lInf) at (-90:3.0) {Infrastructure};
\node[axislab, fill=axReas!18,   draw=axReas]   (lRea) at (-150:3.5){Reasoning Mode};
\node[axislab, fill=axEval!18,   draw=axEval]   (lEva) at ( 150:3.5){Evaluation Methodology};

\foreach \target in {lAss,lOrc,lKno,lInf,lRea,lEva}{
  \draw[->,gray!55, line width=0.35mm] (aob) -- (\target);
}

\node[leaf] (a1) at ($(lAss.north)+(-5.30, 0.45)$) {\textbf{G7} Battery \footnotesize(NASA Li-ion)};
\node[leaf] (a2) at ($(lAss.north)+(-1.75, 0.45)$) {\textbf{G8} Transformer \footnotesize(IEC std.)};
\node[leaf] (a3) at ($(lAss.north)+( 1.75, 0.45)$) {\textbf{G30} PHMForge \footnotesize(8 classes)};
\node[leaf] (a4) at ($(lAss.north)+( 5.30, 0.45)$) {\textbf{G23} Visual Inspect \footnotesize(4 cls., MM)};
\draw[axAsset, line width=0.3mm] (lAss.north) -- (a1.south);
\draw[axAsset, line width=0.3mm] (lAss.north) -- (a2.south);
\draw[axAsset, line width=0.3mm] (lAss.north) -- (a3.south);
\draw[axAsset, line width=0.3mm] (lAss.north) -- (a4.south);

\node[leaf] (o1) at ($(lOrc.east)+(2.1, 0.95)$) {\textbf{G5} Supervisor-Specialist};
\node[leaf] (o2) at ($(lOrc.east)+(2.1, 0.30)$) {\textbf{G12} Verified PE + Self-Ask};
\node[leaf] (o3) at ($(lOrc.east)+(2.1,-0.35)$) {\textbf{G30} ReAct / Claude Code};
\draw[axOrch, line width=0.3mm] (lOrc.east) -- (o1.west);
\draw[axOrch, line width=0.3mm] (lOrc.east) -- (o2.west);
\draw[axOrch, line width=0.3mm] (lOrc.east) -- (o3.west);

\node[leaf] (k1) at ($(lKno.east)+(2.1, 0.30)$) {\textbf{G3} Knowledge Plugin MCP};
\node[leaf] (k2) at ($(lKno.east)+(2.1,-0.35)$) {\textbf{G14} Confidence-gated routing};
\node[leaf] (k3) at ($(lKno.east)+(2.1,-1.00)$) {\textbf{G9} Temporal semantic cache};
\draw[axKnow, line width=0.3mm] (lKno.east) -- (k1.west);
\draw[axKnow, line width=0.3mm] (lKno.east) -- (k2.west);
\draw[axKnow, line width=0.3mm] (lKno.east) -- (k3.west);

\node[leaf] (i1) at ($(lInf.south)+(-5.30,-0.45)$) {\textbf{G16} Hedged FMSR dispatch};
\node[leaf] (i2) at ($(lInf.south)+(-1.75,-0.45)$) {\textbf{G19} TTM $\leftrightarrow$ Chronos swap};
\node[leaf] (i3) at ($(lInf.south)+( 1.75,-0.45)$) {\textbf{G27} torch.compile + GPU};
\node[leaf] (i4) at ($(lInf.south)+( 5.30,-0.45)$) {\textbf{G7} MCP-batching cache};
\draw[axInfra, line width=0.3mm] (lInf.south) -- (i1.north);
\draw[axInfra, line width=0.3mm] (lInf.south) -- (i2.north);
\draw[axInfra, line width=0.3mm] (lInf.south) -- (i3.north);
\draw[axInfra, line width=0.3mm] (lInf.south) -- (i4.north);

\node[leaf] (r1) at ($(lRea.west)+(-2.1, 0.30)$) {\textbf{G21} Reasoning on/off};
\node[leaf] (r2) at ($(lRea.west)+(-2.1,-0.35)$) {\textbf{G20} QLoRA tool internalize};
\node[leaf] (r3) at ($(lRea.west)+(-2.1,-1.00)$) {\textbf{G14} Adaptive routing};
\draw[axReas, line width=0.3mm] (lRea.west) -- (r1.east);
\draw[axReas, line width=0.3mm] (lRea.west) -- (r2.east);
\draw[axReas, line width=0.3mm] (lRea.west) -- (r3.east);

\node[leaf] (e1) at ($(lEva.west)+(-2.1, 0.95)$) {\textbf{G8} Auto Scenario-Gen};
\node[leaf] (e2) at ($(lEva.west)+(-2.1, 0.30)$) {\textbf{G12} Protocol $\times$ Orchestration};
\node[leaf] (e3) at ($(lEva.west)+(-2.1,-0.35)$) {\textbf{G30} MCP-vs-RAG ablation};
\draw[axEval, line width=0.3mm] (lEva.west) -- (e1.east);
\draw[axEval, line width=0.3mm] (lEva.west) -- (e2.east);
\draw[axEval, line width=0.3mm] (lEva.west) -- (e3.east);
\end{tikzpicture}
\caption{Fourteen extensions of one MCP-based industrial-agent benchmark along six axes. Group identifiers index Section~\ref{sec:triangulation}.}
\label{fig:extension-map}
\end{figure*}

\section{Introduction}

The evaluation of LLM agents has outgrown its leaderboards. Agents
today plan, call tools, reuse artifacts across turns, and coordinate
with other agents, yet they are ranked by a small number of aggregate
scores inherited from single-shot benchmarks. The result is rank
instability. In a recent 149-team agentic competition
\citep{patel2026codsretro}, the Spearman correlation between
public-leaderboard rankings and hidden-evaluation rankings was
$\rho=-0.13$ on the execution track ($n=13$, statistically
indistinguishable from zero) and $\rho=0.69$ on the planning track
($n=20$), with the planning track's upper $95\%$ confidence bound
falling at the falsification threshold we propose. Aggregate scores do
not predict what an operator would observe in deployment.

Two streams of work have responded to this gap. The first is a wave
of agent benchmarks
\citep{jimenez2023swebench,yao2024taubench,meta2025are,luo2025mcpuniverse,wang2026mcpbench,patel2025assetopsbench,bandel2026exgentic};
each surfaces a different facet of trajectory-level evaluation, but
each still ranks configurations by aggregate score. The second is a
longer-running NLP critique of single-score ranking
\citep{ethayarajh2020utility,bowman2021will,raji2021everything},
multi-dimensional frameworks
\citep{liang2022helm,kiela2021dynabench}, and behavioral testing
\citep{ribeiro2020checklist}. HELM broadens the dimensions a
single-shot model is measured on, but agents introduce orthogonal
axes (orchestration, multi-turn artifact reuse, tool-call hygiene,
judge-independent verification) that HELM does not score; recent
calls to treat interactive evaluation as a design science
\citep{xuan2026interactive} name the structural change without
proposing a ranking criterion.

The contribution of this paper is that ranking criterion: \textit{predictive
validity}, the correlation between in-sample and out-of-sample rank,
paired with a twelve-tier measurement apparatus that exposes the
dimensions current leaderboards collapse. We study the problem in
industrial asset operations and maintenance, where the stakes of
mis-evaluation are concrete (operator time, capital, safety) and where
fourteen recent implementation studies extending one MCP-based benchmark
have surfaced complementary failure modes the original benchmark's
headline metric does not see.

We argue for a particular position:

\begin{position}
\textbf{Position.} Aggregate-score leaderboards systematically
underspecify the dimensions on which deployed LLM agents are evaluated.
The field would benefit from a multi-tier measurement apparatus ranked
by predictive validity rather than in-sample mean.
\end{position}

The position rests on three supporting claims, each developed below:

\begin{enumerate}\setlength\itemsep{0.2em}
  \item Existing benchmarks measure overlapping subsets of a larger
    measurement space; their findings, taken together, suggest the larger
    space is non-redundant.
  \item Within a single industrial domain, parallel implementations by
    independent teams converge on evaluation dimensions that no single
    benchmark surfaces.
  \item Predictive validity; how in-sample rankings forecast
    out-of-sample rankings; is a more useful ranking criterion than
    in-sample mean for deployment decisions.
\end{enumerate}

This paper draws on three kinds of evidence. First, seven recent
benchmark and deployed-system papers (AssetOpsBench, MCP-Bench,
MCP-Universe, ARE/Gaia2, TaskBench \citep{shen2024taskbench},
Exgentic, and the Condition Insight industrial CMMS framework
\citep{odonncha2026condition}) whose distinctive metrics we consolidate.
Second, fourteen parallel implementation studies (See Figure \ref{fig:extension-map}) of MCP-based agentic
systems in industrial asset operations and maintenance, each producing
complementary failure-mode observations
\citep{li2025multiturn,li2025profiling,li2025skillsknowledge,mazeeva2025skillknowledge,merchant2025temporal,gowda2025agentopsbench,kumar2025transformer,group202025qlora,li2025phmforge,bhandari2025smartgridbench,jebbouri2025fmsrbottleneck,go2025tsfmprofile,vinod2025tsfmoptim}.
Third, cross-paper triangulation: findings observed independently by
multiple teams under different architectures provide stronger inference
than any single empirical study can.

\paragraph{Contributions.} (1) A synthesis of evaluation dimensions
across seven source benchmarks, organized into a twelve-tier measurement
apparatus; (2) a position argument that aggregate-score leaderboards
underspecify deployment-relevant evaluation; (3) three concrete
predictive-validity criteria (held-out, cross-subset, adversarial) for
testing whether the position holds; (4) a predictive-validity leaderboard
schema with implementation guidance; (5) a falsifiable research agenda
for empirical validation, with explicit acknowledgment of what we have
not yet tested.

\paragraph{Benchmark choice.} We anchor the synthesis on
AssetOpsBench \citep{patel2025assetopsbench} because, at time of
writing, it satisfies four criteria that distinguish a serious
deployment-oriented agent benchmark on the public information
available: (i) community traction (more than 1{,}600 GitHub stars and
230 forks); (ii) cross-venue uptake (accepted publications at EMNLP
2025, NeurIPS 2025, and AAAI 2026); (iii) demonstrated extensibility
(used as the substrate of the 149-team CODS-2025 competition
\citep{patel2026codsretro} and of the fourteen implementation studies
aggregated in this paper, each adding scenarios, asset classes, or
architectural variants); and (iv) full public availability of the
benchmark scenarios, MCP server tooling, and Docker artifacts, which
is what made deep-dive replication tractable for those teams.

\paragraph{Scope.} This is a position paper grounded in synthesis;
we do not run new controlled experiments. The fourteen implementation
studies were selected for currency, depth of extension
along a single axis (each study modifies one architectural variable
end-to-end), and coverage of complementary tiers of the apparatus we
propose. We therefore frame the evidence as convergent architectural
sensitivity rather than independent triangulation: the same benchmark
surfaces different failure modes under different architectural
choices. We commit to the falsification conditions in
Section~\ref{sec:falsification}.

\section{The Argument}

The position rests on three structural critiques of aggregate-score
leaderboards; the transferability assumption, judge reflexivity, and
the circularity through which evaluation constitutes the capability it
measures \citep{kalaitzidis2026evaltrap}; developed below.

\subsection{Aggregate Scores Collapse Orthogonal Dimensions}

The core argument: a Pass$^{1}$ score of $0.75$ can be achieved by many
qualitatively different configurations; one that is reasoning-heavy and
cost-expensive, one that is retrieval-rich and latency-bound, one that is
tool-hygiene-fragile but artifact-reuse-efficient. Aggregate scores treat
these as equivalent; deployment treats them as not. Three concrete cases
of aggregation hiding qualitatively-distinct behavior, each later
re-introduced as a sensitivity case in Section~\ref{sec:triangulation}:

\begin{itemize}\setlength\itemsep{0.2em}
  \item \textbf{Per-rubric reasoning sensitivity.} Reasoning-on vs.\
    reasoning-off configurations score similarly on overall rubric mean
    but differ by 31 percentage points on clarity-specific scoring, while
    data-retrieval and agent-sequence dimensions are unchanged
    \citep{li2025profiling}.
  \item \textbf{Multi-turn artifact reuse.} Plan-Execute and
    Supervisor-Specialist architectures score similarly on single-turn
    Pass$^{1}$ but differ by $4.2\times$ on turn-2-to-5 latency due to
    cross-turn artifact reuse; a dimension invisible in single-turn
    benchmarks \citep{li2025multiturn}.
  \item \textbf{Retrieval-strategy trade-off.} Single-pass RAG and
    agentic multi-hop retrieval show $50$--$68\%$ vs.\ $\sim$$90\%$
    accuracy with $4.5\times$--$10\times$ token inflation
    \citep{li2025skillsknowledge}. Neither dominates; the right choice
    depends on deployment constraints aggregate scores do not surface.
\end{itemize}

\subsection{LLM-as-Judge Measurement is Reflexive}

Most leaderboards depend on LLM-as-judge scoring
\citep{zheng2023mtbench}, which is itself a measurement instrument with
model-specific biases. As judge models evolve, ranking shifts; as judge
prompts are adjusted, scores move. The leaderboard risks measuring its
own judge as much as the systems it evaluates.

Two independent efforts demonstrate that judge-independent measurement is
feasible. Condition Insight \citep{odonncha2026condition} introduces the
Condition Agreement Rate (CAR) by comparing LLM-assigned
classifications to a parallel rule-based pipeline; reported CAR moves
from $0.68$ to $0.91$ under a constrained-prompting design, a 20pp
improvement attributable to prompting rather than to backbone-model
choice. ARE/Gaia2 \citep{meta2025are} checks executed trajectories
against human-annotated oracle DAGs with hard, causality, and timing
checks that require no LLM evaluation; the reported verifier achieves
$0.99$ precision and $0.95$ recall against 450 hand-labeled
trajectories. A third data point sharpens the concern: in the PHMForge
benchmark \citep{li2025phmforge}, LLM-as-judge inter-rater reliability
on a 30-scenario stratified sample reaches only Krippendorff
$\alpha=0.61$, well below the human--human range
$\alpha\in[0.74,0.82]$ on the same sample. The judge is itself a
weaker rater than the experts it is meant to surrogate; a finding that
generalizes the methodological point: a leaderboard with no
judge-independent component has no anchor against which to detect judge
drift.

\subsection{Out-of-Distribution Behavior is the Deployment Question}

Deployed systems do not encounter the training set or the leaderboard
set. They encounter scenarios that are either (i) distributionally
similar to held-out cases, (ii) distributionally distinct (cross-domain
transfer), or (iii) adversarially perturbed by user phrasing. In-sample
mean predicts none of these directly.

Exgentic's analysis observed cross-benchmark rank correlations of
$0.32$--$0.85$ across six heterogeneous benchmarks
\citep{bandel2026exgentic}, concluding that ``current architectures do
not achieve robust generalization but instead optimize for specific task
distributions.'' We extend this concern: even within a single domain,
leaderboards should test rank stability under distributional shifts.

Anchoring this in earlier NLP work: \citet{ethayarajh2020utility} note
that leaderboard ranks reflect a user's utility function only by
accident; \citet{dehghani2021benchmark} document the ``benchmark lottery''
in which subsample choice reorders rankings; \citet{recht2019imagenet}
provide the classical ImageNet result that ranks fall under modest
distributional shift. A recent agent-side data point sharpens the
concern: \citet{patel2026codsretro} analyze the CODS-2025 AssetOpsBench
challenge (149 teams). Public--hidden Spearman is $\rho=0.69$ for
planning ($n=20$, robustly positive) and $\rho=-0.13$ for execution
($n=13$, $p=0.71$, statistically indistinguishable from zero). With
$n=13$, a $95\%$ bootstrap CI on the execution correlation spans
roughly $[-0.64, +0.45]$; wide enough that we treat the execution
result as suggestive only. The planning track is where the evidence
is robust: rankings do correlate, but mean private scores still fall
$11.3$ points below public, and public scores saturate
(8 unique values across 20 teams). The deployment-relevant question is
not ``what is the mean,'' but ``does this ranking transfer.''

\section{The Synthesis}

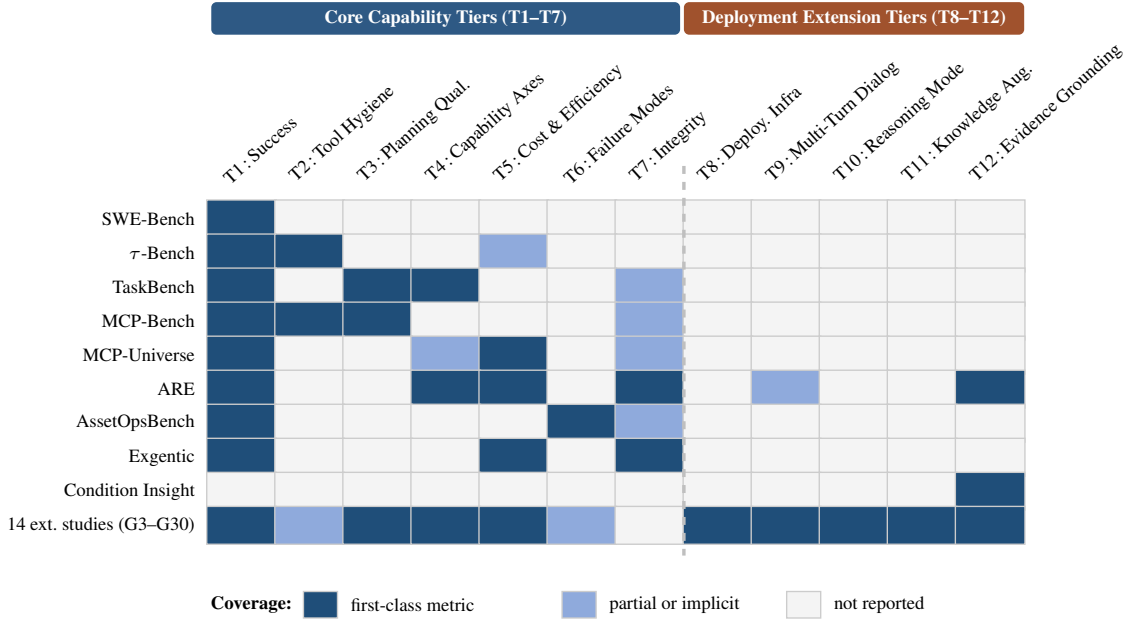
\begin{figure*}[!t]
\centering
\definecolor{covFull}{HTML}{1F4E79}
\definecolor{covPart}{HTML}{8FAADC}
\definecolor{covEmp}{HTML}{F4F4F4}
\definecolor{coreHd}{HTML}{2E5984}
\definecolor{extHd}{HTML}{A0522D}
\begin{tikzpicture}[
  bhd/.style={font=\scriptsize, anchor=south west, rotate=42, inner sep=2pt},
  rlbl/.style={font=\scriptsize, anchor=east, inner sep=2pt},
  cell/.style={rectangle, draw=gray!40, line width=0.18mm,
               minimum width=0.92cm, minimum height=0.5cm, inner sep=0},
  full/.style={cell, fill=covFull},
  part/.style={cell, fill=covPart},
  emp/.style ={cell, fill=covEmp},
  banner/.style={rectangle, rounded corners=2pt, line width=0.2mm,
                font=\bfseries\scriptsize, anchor=center, inner sep=3pt, text=white}
]
\def\cs{0.90}
\def\rs{0.45}
\foreach \i/\name in {1/Success, 2/Tool Hygiene, 3/Planning Qual., 4/Capability Axes, 5/Cost \& Efficiency, 6/Failure Modes, 7/Integrity, 8/Deploy. Infra, 9/Multi-Turn Dialog, 10/Reasoning Mode, 11/Knowledge Aug., 12/Evidence Grounding}{
  \node[bhd] at (\i*\cs - 0.15, 0.34) {T\i\,:\,\name};
}
\node[banner, fill=coreHd, minimum width=6.2cm] at (4.0*\cs, 2.65) {Core Capability Tiers (T1--T7)};
\node[banner, fill=extHd, minimum width=4.5cm] at (10.0*\cs, 2.65) {Deployment Extension Tiers (T8--T12)};

\foreach \r/\lbl in {0/SWE-Bench, 1/$\tau$-Bench, 2/TaskBench, 3/MCP-Bench, 4/MCP-Universe, 5/ARE/Gaia2, 6/AssetOpsBench, 7/Exgentic, 8/Condition Insight, 9/{14 ext.\ studies (G3--G30)}}{
  \node[rlbl] at (0.4*\cs, \r*-\rs) {\lbl};
}
\node[full] at (1*\cs, 0*-\rs) {};
\foreach \c in {2,...,12}{\node[emp] at (\c*\cs, 0*-\rs) {};}
\node[full] at (1*\cs, 1*-\rs) {};
\node[full] at (2*\cs, 1*-\rs) {};
\node[emp]  at (3*\cs, 1*-\rs) {};
\node[emp]  at (4*\cs, 1*-\rs) {};
\node[part] at (5*\cs, 1*-\rs) {};
\foreach \c in {6,...,12}{\node[emp] at (\c*\cs, 1*-\rs) {};}
\node[full] at (1*\cs, 2*-\rs) {};
\node[emp]  at (2*\cs, 2*-\rs) {};
\node[full] at (3*\cs, 2*-\rs) {};
\node[full] at (4*\cs, 2*-\rs) {};
\node[emp]  at (5*\cs, 2*-\rs) {};
\node[emp]  at (6*\cs, 2*-\rs) {};
\node[part] at (7*\cs, 2*-\rs) {};
\foreach \c in {8,...,12}{\node[emp] at (\c*\cs, 2*-\rs) {};}
\node[full] at (1*\cs, 3*-\rs) {};
\node[full] at (2*\cs, 3*-\rs) {};
\node[full] at (3*\cs, 3*-\rs) {};
\node[emp]  at (4*\cs, 3*-\rs) {};
\node[emp]  at (5*\cs, 3*-\rs) {};
\node[emp]  at (6*\cs, 3*-\rs) {};
\node[part] at (7*\cs, 3*-\rs) {};
\foreach \c in {8,...,12}{\node[emp] at (\c*\cs, 3*-\rs) {};}
\node[full] at (1*\cs, 4*-\rs) {};
\node[emp]  at (2*\cs, 4*-\rs) {};
\node[emp]  at (3*\cs, 4*-\rs) {};
\node[part] at (4*\cs, 4*-\rs) {};
\node[full] at (5*\cs, 4*-\rs) {};
\node[emp]  at (6*\cs, 4*-\rs) {};
\node[part] at (7*\cs, 4*-\rs) {};
\foreach \c in {8,...,12}{\node[emp] at (\c*\cs, 4*-\rs) {};}
\node[full] at (1*\cs, 5*-\rs) {};
\node[emp]  at (2*\cs, 5*-\rs) {};
\node[emp]  at (3*\cs, 5*-\rs) {};
\node[full] at (4*\cs, 5*-\rs) {};
\node[full] at (5*\cs, 5*-\rs) {};
\node[emp]  at (6*\cs, 5*-\rs) {};
\node[full] at (7*\cs, 5*-\rs) {};
\node[emp]  at (8*\cs, 5*-\rs) {};
\node[part] at (9*\cs, 5*-\rs) {};
\node[emp]  at (10*\cs, 5*-\rs) {};
\node[emp]  at (11*\cs, 5*-\rs) {};
\node[full] at (12*\cs, 5*-\rs) {};
\node[full] at (1*\cs, 6*-\rs) {};
\node[emp]  at (2*\cs, 6*-\rs) {};
\node[emp]  at (3*\cs, 6*-\rs) {};
\node[emp]  at (4*\cs, 6*-\rs) {};
\node[emp]  at (5*\cs, 6*-\rs) {};
\node[full] at (6*\cs, 6*-\rs) {};
\node[part] at (7*\cs, 6*-\rs) {};
\foreach \c in {8,...,12}{\node[emp] at (\c*\cs, 6*-\rs) {};}
\node[full] at (1*\cs, 7*-\rs) {};
\node[emp]  at (2*\cs, 7*-\rs) {};
\node[emp]  at (3*\cs, 7*-\rs) {};
\node[emp]  at (4*\cs, 7*-\rs) {};
\node[full] at (5*\cs, 7*-\rs) {};
\node[emp]  at (6*\cs, 7*-\rs) {};
\node[full] at (7*\cs, 7*-\rs) {};
\foreach \c in {8,...,12}{\node[emp] at (\c*\cs, 7*-\rs) {};}
\foreach \c in {1,...,11}{\node[emp] at (\c*\cs, 8*-\rs) {};}
\node[full] at (12*\cs, 8*-\rs) {};
\node[full] at (1*\cs, 9*-\rs) {};
\node[part] at (2*\cs, 9*-\rs) {};
\node[full] at (3*\cs, 9*-\rs) {};
\node[full] at (4*\cs, 9*-\rs) {};
\node[full] at (5*\cs, 9*-\rs) {};
\node[part] at (6*\cs, 9*-\rs) {};
\node[emp]  at (7*\cs, 9*-\rs) {};
\node[full] at (8*\cs, 9*-\rs) {};
\node[full] at (9*\cs, 9*-\rs) {};
\node[full] at (10*\cs, 9*-\rs) {};
\node[full] at (11*\cs, 9*-\rs) {};
\node[full] at (12*\cs, 9*-\rs) {};

\draw[gray!50, dashed, line width=0.45mm] (7.5*\cs, 0.7) -- (7.5*\cs, -9*\rs-0.4);

\node[font=\scriptsize\bfseries, anchor=west] at (0.4*\cs, -9*\rs - 1.05) {Coverage:};
\node[full, anchor=west, minimum width=0.45cm, minimum height=0.35cm] at (1.9*\cs, -9*\rs - 1.05) {};
\node[font=\scriptsize, anchor=west] at (1.9*\cs + 0.5, -9*\rs - 1.05) {first-class metric};
\node[part, anchor=west, minimum width=0.45cm, minimum height=0.35cm] at (5.7*\cs, -9*\rs - 1.05) {};
\node[font=\scriptsize, anchor=west] at (5.7*\cs + 0.5, -9*\rs - 1.05) {partial or implicit};
\node[emp, anchor=west, minimum width=0.45cm, minimum height=0.35cm] at (9.0*\cs, -9*\rs - 1.05) {};
\node[font=\scriptsize, anchor=west] at (9.0*\cs + 0.5, -9*\rs - 1.05) {not reported};
\end{tikzpicture}
\caption{Benchmark $\times$ tier coverage. No single prior benchmark reports more than four or five tiers; the deployment-extension tiers (T8--T12) are absent from almost all of them.}
\label{fig:twelve-tiers}
\end{figure*}

The OOD-rank-instability argument above asks \textit{whether} ranks
transfer. The next question is \textit{what to measure} so that they
can. The twelve tiers in Figure~\ref{fig:twelve-tiers} are not invented
but consolidated from the distinctive contributions of seven source
benchmarks plus fourteen parallel implementation studies. Each tier
corresponds to a measurement dimension that at least one prior work
surfaces and that we argue is non-redundant with the others; the
``roughly twelve'' count and the orthogonality claim are working
hypotheses, tested empirically in the research agenda not in the
present paper.

The seven core tiers (T1--T7) consolidate metrics from prior
benchmarks: pass-rate floors (T1), tool-call hygiene (T2),
planning-process quality (T3), capability axes (T4), cost-efficiency
Pareto (T5), failure-mode taxonomies (T6), and reproducibility (T7).
The five deployment-extension tiers (T8--T12) are surfaced by the
fourteen implementation studies of Section~\ref{sec:triangulation}:
deployment infrastructure (T8), multi-turn dialog (T9), reasoning-mode
adaptivity (T10), knowledge augmentation (T11), and evidence
grounding with judge-independent verification (T12). Appendix
\ref{appx:tier-defs} gives the metric-level definition for each tier
along with the prior work that surfaces it.

\begin{position}
\textit{The twelve-tier apparatus is not a proposal for a maximally
comprehensive leaderboard. It is a claim that the measurement space of
deployed LLM agents has roughly twelve orthogonal-or-near-orthogonal
dimensions, that no current leaderboard reports more than four or five of
them, and that the gap between current practice and the underlying
measurement space drives the systematic over-trusting of benchmark
rankings that motivates this position.}
\end{position}

\section{Predictive Validity as Ranking Criterion}
\label{sec:predictive-validity}

The twelve tiers say \textit{what} to measure. The next question is
\textit{how} to use those measurements to rank configurations. A
leaderboard's purpose is to inform deployment decisions, and
deployment decisions depend on out-of-sample performance, not in-sample
performance. The right ranking criterion is therefore predictive
validity: the correlation between in-sample rank and out-of-sample
rank, not in-sample mean.

\subsection{Three Operationalizations of OOD Shift}

\paragraph{Criterion A: Held-Out Scenarios (mild shift).} Stratified
random split of an existing benchmark, preserving the joint distribution
of subset and category across the split. Tests whether the leaderboard
ranking on a sample predicts ranking on the full population. Weakest
test; passes are uninformative, failures are damning.

\paragraph{Criterion B: Cross-Subset Transfer (moderate shift).} Rank on
$k-1$ subsets, test on the held-out subset; rotate across all subsets.
For AssetOpsBench's six subsets (HVAC chillers, compressor, hydraulic
pump, FMSA, PHM, rule logic), this produces a $6\times 6$ rank-stability
matrix. The most realistic test: ``you ranked agents on chillers; will
the ranking transfer to hydraulic pumps?''

\paragraph{Criterion C: Adversarial Perturbation (strongest shift).}
Semantically-equivalent paraphrases of base scenarios across four
classes: paraphrase (rewrite query preserving intent),
identifier renaming (e.g., Chiller~6 $\rightarrow$ Unit-CHX-06),
time-window shifting (``last week'' $\rightarrow$ ``two weeks
ago to one week ago''), and distractor injection (append
irrelevant operational context). A configuration that genuinely solved
the underlying task should perform equivalently across base and
perturbed versions.

\subsection{The Predictive-Validity Score}

A composite ranking score that combines mean performance with
out-of-sample reliability:
\begin{equation*}
\text{PV}(c) = \alpha\, \bar{Y}_c - \beta\, \sigma_{Y_c,\text{OOD}} - \gamma\, \text{IQR}(Y_c)
\end{equation*}
where $\bar{Y}_c$ is in-sample mean, $\sigma_{Y_c,\text{OOD}}$ is
cross-OOD-criterion standard deviation of rank position, and
$\text{IQR}(Y_c)$ is interquartile range of per-scenario scores. The
weights $\alpha,\beta,\gamma$ are fit on Criterion-A holdouts to maximize
Spearman correlation between PV rank and Criterion-B/C ranks. We do not
finalize the weights or the functional form in this position paper. The
score is a proposal; the empirical study to fit and validate it is part
of the research agenda in Section~\ref{sec:agenda}.

\subsection{Falsification of the Position}
\label{sec:falsification}

For our position to be considered well-supported under empirical
validation, the following conditions should hold:

\begin{itemize}\setlength\itemsep{0.2em}
  \item Spearman rank correlation between in-sample and OOD rankings
    $\rho<0.85$ across at least two of three OOD criteria (else
    generalization is high and our concern is moot).
  \item Top-3 in-sample configurations leave top-5 on out-of-sample in
    $\geq 10\%$ of holdout splits (else top-3 deployment recommendations
    would not change).
  \item Mean-vs.-OOD-variance correlation $\rho_{\text{Pearson}}>0.2$
    (else high-performing configurations are not disproportionately
    unstable).
  \item PV-ranking top-10 differs from mean-ranking top-10 with Jaccard
    overlap $<0.85$ (else the proposed methodology offers no different
    deployment guidance).
\end{itemize}

We commit to publishing the position as refuted if these conditions fail
under controlled experiment. The first condition is already partially
supported: \citet{patel2026codsretro} report execution-track public--private
Spearman $\rho=-0.13$, well below the $0.85$ threshold, on the same
benchmark family this paper studies. The remaining conditions require
the controlled study outlined in Section~\ref{sec:agenda}.

\begin{figure*}[!t]
\centering
\definecolor{barAsset}{HTML}{C0392B}
\definecolor{barOrch}{HTML}{2E86AB}
\definecolor{barKnow}{HTML}{1E8449}
\definecolor{barInfra}{HTML}{D68910}
\definecolor{barReas}{HTML}{7D3C98}
\definecolor{barEval}{HTML}{B9591E}
\begin{tikzpicture}[
  rowlabel/.style={font=\scriptsize, anchor=east, inner sep=2pt},
  annot/.style ={font=\tiny\itshape, text=gray!55!black, anchor=west, inner sep=2pt},
  axislabel/.style={font=\scriptsize, anchor=north},
  tick/.style={font=\tiny, text=gray!55!black, anchor=north, inner sep=1pt},
  bar/.style={rectangle, draw=none, anchor=west, minimum height=0.36cm, inner sep=0}
]
\def\scl{1.80}  
\def\xz{0}     

\foreach \v/\lab in {0/1$\times$, 0.477/3$\times$, 1/10$\times$, 1.477/30$\times$, 2/100$\times$, 2.477/300$\times$, 3/1000$\times$, 3.544/3500$\times$}{
  \draw[gray!40, dashed, line width=0.15mm] (\v*\scl, 0.3) -- (\v*\scl, -7.6);
  \node[tick] at (\v*\scl, -7.65) {\lab};
}


\def\rH{0.50}
\def\bars{%
  {0/G7 disk-cache predict / 3.544/barAsset/G7 disk cache 3500$\times$ (predict-only)/quality preserved (8/11 pass)},
  {1/G16 Hedged FMSR / 1.556/barInfra/G16 Hedged FMSR 36$\times$ /quality 0.660$\to$0.675 (INT4 3B vs 70B)},
  {2/G8 ScenarioGen / 0.903/barEval/G8 Scenario-Gen 8$\times$ /quality preserved (74.2 vs 73.8)},
  {3/G20 QLoRA tokens / 0.759/barReas/G20 QLoRA 5.7$\times$ token reduction/AT-F1 +0.18 (0.47$\to$0.65)},
  {4/G7 MCP-batched fleet / 0.783/barAsset/G7 MCP-batched fleet 6.06$\times$/quality preserved},
  {5/G5 Multi-Turn 2-to-5 / 0.623/barOrch/G5 Multi-Turn 4.2$\times$ (turn 2-5 vs 1)/completion +30pp},
  {6/G9 Temporal Cache / 0.541/barKnow/G9 Temporal cache 3.48$\times$ end-to-end/F1 cap 0.64 (cache-validity)},
  {7/G27 TSFM Optim / 0.519/barInfra/G27 torch.compile + GPU 3.3$\times$/quality preserved},
  {8/G14 Conf-Gate / 0.369/barReas/G14 Confidence-gate 2.34$\times$ overall correct/hallucination $-58$pp},
  {9/G12 Verified PE+SA / 0.155/barEval/G12 Verified PE + Self-Ask 1.43$\times$ judge mean/+8.4pp pass / +3.2$\times$ latency},
  {10/G21 Reasoning-on / 0.103/barReas/G21 Reasoning-on 1.26$\times$ completion/clarity +31pp / +21.5\% latency},
  {11/G3 KP vs RAG / 0.342/barKnow/G3 Knowledge Plugin 2.2$\times$ accuracy ratio/+22pp / 4.5-10$\times$ tokens},
  {12/G23 VI Quant / 0.299/barAsset/G23 Visual Inspect AWQ-domain 1.99$\times$/+34pp judge pass (0.48$\to$0.82)}%
}

\foreach \i/\name/\lv/\cl/\lbl/\an in \bars {
  \pgfmathsetmacro\yp{-\i*\rH}
  \node[rowlabel] at (-0.15, \yp) {\textbf{\lbl}};
  \draw[\cl, line width=0.5mm, fill=\cl, fill opacity=0.85] (\xz, \yp-0.10) rectangle (\lv*\scl, \yp+0.10);
  \node[annot] at (\lv*\scl + 0.1, \yp) {\an};
}

\node[font=\tiny\bfseries, anchor=west] at (-3.5, -7.0) {Axis colors:};
\foreach \i/\name/\cl in {0/Asset/barAsset, 1/Orch./barOrch, 2/Know./barKnow, 3/Infra/barInfra, 4/Reas./barReas, 5/Eval/barEval}{
  \fill[\cl] (-2.3 + \i*1.0, -7.10) rectangle (-2.1 + \i*1.0, -6.96);
  \node[font=\tiny, anchor=west] at (-2.05 + \i*1.0, -7.03) {\name};
}
\node[axislabel] at (3*\scl, -8.2) {\textbf{Speedup or improvement ratio (log scale)}};
\end{tikzpicture}
\caption{Headline improvement per team vs.\ own baseline, log-scale, colored by extension axis. G9, G19, G30 omitted (non-magnitude findings; see Table~\ref{tab:per-study-setup}).}
\label{fig:magnitude}
\end{figure*}
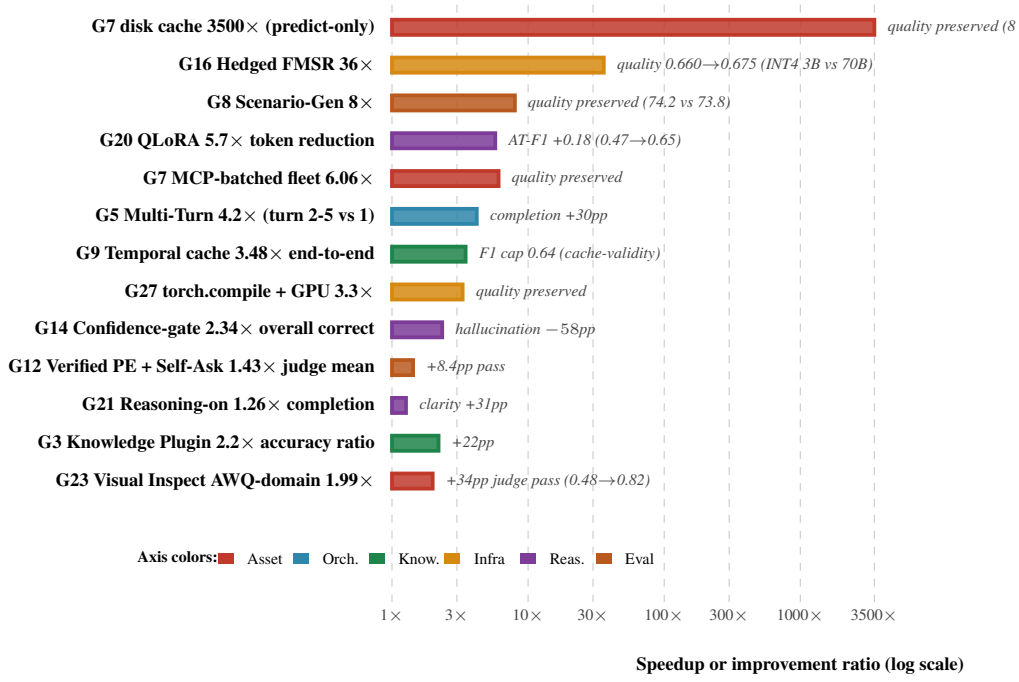

Figure~\ref{fig:magnitude} shows the headline improvement each team
reports against its own baseline. Bars are colored by primary extension
axis to match Figure~\ref{fig:extension-map}. The horizontal axis is
logarithmic because the range spans roughly $1.5\times$ to
$3500\times$. Annotations state what was held constant or co-improved
(``quality preserved'', accuracy gain, etc.). Three groups (G9, G19,
G30) are not plotted because their headline findings are not magnitudes
(F1 ceiling, workflow-dependent backbone swap, substrate ablation); see
Appendix~\ref{appx:per-study} for those.

\section{Convergent Architectural Sensitivity}
\label{sec:triangulation}

Where Section~\ref{sec:predictive-validity} prescribes a ranking
criterion, this section asks whether the underlying measurements are
even being collected. We report findings from fourteen implementation
studies of MCP-based industrial agents, grouped by the extension axis
each touches most directly (matching Figure~\ref{fig:extension-map}).
Figure~\ref{fig:magnitude} previews the magnitude of improvement each
team reports against its own baseline; the cases below explain what
each bar represents and which evaluation dimension it implicates. The
studies share institutional context as noted in Section~1, so we treat
their convergence as architectural sensitivity rather than independent
replication. Variances and confidence intervals are reported where the
source studies provide them.

\subsection{Reasoning mode: cost of extended thinking}

A plan-execute profiling study toggled extended-thinking on a
Gemma-4-26B planner over 40 multi-agent AssetOpsBench scenarios served
through vLLM on an A100 \citep{li2025profiling}. Reasoning-on raised
total latency by $21.5\%$ (15.08\,s to 18.32\,s) and planning latency
by $41.9\%$. The quality gain was non-uniform: clarity improved 31
percentage points (61\% to 92\%) and hallucination dropped 7\,pp
(12\% to 5\%), but data retrieval and agent-sequence correctness were
unchanged. The overall rubric mean hides where reasoning helps and
where it does not; per-rubric reasoning sensitivity belongs in any
leaderboard that reports reasoning-mode results.

\subsection{Knowledge augmentation: RAG versus Knowledge Plugin}

A skills-and-knowledge-plugin study compared single-pass RAG against
an agentic multi-hop Knowledge-Plugin MCP server on ten AssetOpsBench
scenarios \citep{li2025skillsknowledge}. With Llama-4-Maverick-17B,
RAG reached $50$--$68\%$ rubric accuracy at 8.9--20\,s end-to-end,
while the Knowledge Plugin reached ${\sim}90\%$ at 114--146\,s with
$4.5\times$--$10\times$ token inflation. A cross-model split sharpened
the trade-off: Granite-3-8B reached 60\% at 91\,s on the same Knowledge
Plugin pipeline, a weaker model with a faster path and a lower ceiling.
Neither configuration dominates, and a leaderboard reporting only
accuracy hides the deployment regime that selects between them.
Retrieval strategy (single-pass, multi-hop, hybrid) should therefore
be a declared submission property.

\subsection{Evaluation methodology: judge-independent governance}

Two deployed systems and one new benchmark all show that LLM-as-judge
scoring needs an external anchor. Condition Insight, an industrial
CMMS reasoning framework, verifies LLM-assigned condition
classifications against a parallel rule-based pipeline and reports
Condition Agreement Rate (CAR) climbing from $0.68$ to $0.91$ under a
constrained-prompting design, a prompt-level gain rather than a
backbone-model gain \citep{odonncha2026condition}. ARE/Gaia2's DAG
oracle does analogous trajectory-level verification, achieving $0.99$
precision and $0.95$ recall against 450 hand-labeled trajectories
\citep{meta2025are}. PHMForge sharpens the concern from the judge
side: LLM-as-judge inter-rater reliability sits at Krippendorff
$\alpha=0.61$ on a 30-scenario stratified sample, well below the
human-human range $\alpha \in [0.74, 0.82]$ on the same data
\citep{li2025phmforge}. A leaderboard with no judge-independent
component has no anchor against which to detect judge drift.

\subsection{Evaluation methodology: substrate underspecification}

The PHMForge benchmark stages 99 SME-authored prognostics scenarios
across eight industrial asset classes, served through 39
algorithm-grounded MCP tools \citep{li2025phmforge}. It is the only
study in our corpus that replaces, rather than extends, the
plan-execute base, using ReAct and Claude Code instead. The strongest
configuration reached $80.8\%$ pass@1, but three controlled ablations
mattered more than the headline number. Replacing MCP tool execution
with text-RAG over telemetry-equivalent evidence collapsed
remaining-useful-life pass-all-3 from $100\%$ to $20\%$ (5/5 to 1/5)
on the lithium-ion class. Cross-equipment transfer dropped pass rate
from $84.1\%$ on bearings to $42.7\%$ on motors, a $41$-point gap on
the same scenario format. Operator-style fuzzy queries fell from
$80.6\%$ to $48.6\%$ (McNemar $p=0.002$), and withholding domain tools
took completion from $80.8\%$ down to $25\%$. Substrate, asset, and
query-formulation are three deployment-relevant axes invisible to any
fixed-substrate, fixed-asset, fixed-phrasing leaderboard; orchestration
errors dominate failures, and frontier LLMs in the authors' words are
``stronger at calling tools than at planning when to call them.''

\subsection{The Triangulation Claim}

\begin{claim}
\textit{No one of these findings alone justifies a twelve-tier apparatus.
Taken together, fourteen parallel implementation teams plus seven prior
benchmark papers converged on the same set of measurement dimensions,
each from a distinct architectural starting point. This convergence is
the strongest evidence we offer for the synthesis; not new experiments,
but convergent architectural sensitivity across many.}
\end{claim}

Table~\ref{tab:dimensions} summarizes the mapping from each implementation
study to the four high-level evaluation dimensions the EMNLP theme
foregrounds; robustness, trustworthiness, generalization, and
longitudinal/drift behavior. All four dimensions receive convergent
evidence from multiple independent teams.

\begin{figure}[t]
\centering
\definecolor{hmFill}{HTML}{2E75B6}
\definecolor{hmEmpty}{HTML}{F4F4F4}
\definecolor{hmEdge}{HTML}{8FAADC}
\begin{tikzpicture}[
  rowlbl/.style={font=\scriptsize, anchor=east, inner sep=2pt},
  collbl/.style={font=\footnotesize\bfseries, anchor=south, inner sep=2pt},
  hit/.style ={rectangle, fill=hmFill,  draw=hmEdge!50!black, line width=0.18mm, minimum width=1.05cm, minimum height=0.52cm, inner sep=0},
  miss/.style={rectangle, fill=hmEmpty, draw=gray!40,         line width=0.18mm, minimum width=1.05cm, minimum height=0.52cm, inner sep=0},
  num/.style ={font=\scriptsize, text=white,         anchor=center, inner sep=0},
  numM/.style={font=\tiny, text=gray!55!black, anchor=center, inner sep=0}
]
\def\colS{1.20}
\node[collbl] at (1*\colS, 0.42) {Rob.};
\node[collbl] at (2*\colS, 0.42) {Trust.};
\node[collbl] at (3*\colS, 0.42) {Gen.};
\node[collbl] at (4*\colS, 0.42) {Long.};

\def\rowH{0.58}
\node[rowlbl] at (0, 0*-\rowH) {Profiling (G21)};
\node[miss] at (1*\colS, 0*-\rowH) {};
\node[hit]  at (2*\colS, 0*-\rowH) {}; \node[num] at (2*\colS, 0*-\rowH) {31pp};
\node[miss] at (3*\colS, 0*-\rowH) {};
\node[miss] at (4*\colS, 0*-\rowH) {};
\node[rowlbl] at (0, 1*-\rowH) {Multi-Turn (G5)};
\node[hit]  at (1*\colS, 1*-\rowH) {}; \node[num] at (1*\colS, 1*-\rowH) {4.2$\times$};
\node[hit]  at (2*\colS, 1*-\rowH) {}; \node[num] at (2*\colS, 1*-\rowH) {$-69$};
\node[miss] at (3*\colS, 1*-\rowH) {};
\node[miss] at (4*\colS, 1*-\rowH) {};
\node[rowlbl] at (0, 2*-\rowH) {Skills+KP (G3)};
\node[miss] at (1*\colS, 2*-\rowH) {};
\node[hit]  at (2*\colS, 2*-\rowH) {}; \node[num] at (2*\colS, 2*-\rowH) {+22};
\node[miss] at (3*\colS, 2*-\rowH) {};
\node[miss] at (4*\colS, 2*-\rowH) {};
\node[rowlbl] at (0, 3*-\rowH) {Conf.~Gate (G14)};
\node[miss] at (1*\colS, 3*-\rowH) {};
\node[hit]  at (2*\colS, 3*-\rowH) {}; \node[num] at (2*\colS, 3*-\rowH) {$-58$};
\node[miss] at (3*\colS, 3*-\rowH) {};
\node[miss] at (4*\colS, 3*-\rowH) {};
\node[rowlbl] at (0, 4*-\rowH) {Temp.~Cache (G9)};
\node[hit]  at (1*\colS, 4*-\rowH) {}; \node[num] at (1*\colS, 4*-\rowH) {3.5$\times$};
\node[hit]  at (2*\colS, 4*-\rowH) {}; \node[num] at (2*\colS, 4*-\rowH) {0.64};
\node[miss] at (3*\colS, 4*-\rowH) {};
\node[miss] at (4*\colS, 4*-\rowH) {};
\node[rowlbl] at (0, 5*-\rowH) {QLoRA (G20)};
\node[hit]  at (1*\colS, 5*-\rowH) {}; \node[num] at (1*\colS, 5*-\rowH) {$-83\%$};
\node[miss] at (2*\colS, 5*-\rowH) {};
\node[miss] at (3*\colS, 5*-\rowH) {};
\node[hit]  at (4*\colS, 5*-\rowH) {}; \node[num] at (4*\colS, 5*-\rowH) {$-19$};
\node[rowlbl] at (0, 6*-\rowH) {Transformer (G8)};
\node[miss] at (1*\colS, 6*-\rowH) {};
\node[hit]  at (2*\colS, 6*-\rowH) {}; \node[num] at (2*\colS, 6*-\rowH) {74.2};
\node[hit]  at (3*\colS, 6*-\rowH) {}; \node[num] at (3*\colS, 6*-\rowH) {8$\times$};
\node[miss] at (4*\colS, 6*-\rowH) {};
\node[rowlbl] at (0, 7*-\rowH) {Battery (G7)};
\node[hit]  at (1*\colS, 7*-\rowH) {}; \node[num] at (1*\colS, 7*-\rowH) {6$\times$};
\node[miss] at (2*\colS, 7*-\rowH) {};
\node[hit]  at (3*\colS, 7*-\rowH) {}; \node[num] at (3*\colS, 7*-\rowH) {Li-ion};
\node[miss] at (4*\colS, 7*-\rowH) {};
\node[rowlbl] at (0, 8*-\rowH) {PHMForge (G30)};
\node[hit]  at (1*\colS, 8*-\rowH) {}; \node[num] at (1*\colS, 8*-\rowH) {$-41$};
\node[hit]  at (2*\colS, 8*-\rowH) {}; \node[num] at (2*\colS, 8*-\rowH) {0.61};
\node[hit]  at (3*\colS, 8*-\rowH) {}; \node[num] at (3*\colS, 8*-\rowH) {8 cls.};
\node[miss] at (4*\colS, 8*-\rowH) {};
\node[rowlbl] at (0, 9*-\rowH) {SmartGrid (G12)};
\node[hit]  at (1*\colS, 9*-\rowH) {}; \node[num] at (1*\colS, 9*-\rowH) {2.4k};
\node[hit]  at (2*\colS, 9*-\rowH) {}; \node[num] at (2*\colS, 9*-\rowH) {+12};
\node[miss] at (3*\colS, 9*-\rowH) {};
\node[miss] at (4*\colS, 9*-\rowH) {};
\node[rowlbl] at (0,10*-\rowH) {FMSR Disp. (G16)};
\node[hit]  at (1*\colS,10*-\rowH) {}; \node[num] at (1*\colS,10*-\rowH) {36$\times$};
\node[miss] at (2*\colS,10*-\rowH) {};
\node[hit]  at (3*\colS,10*-\rowH) {}; \node[num] at (3*\colS,10*-\rowH) {Wind};
\node[miss] at (4*\colS,10*-\rowH) {};
\node[rowlbl] at (0,11*-\rowH) {TSFM Back. (G19)};
\node[miss] at (1*\colS,11*-\rowH) {};
\node[miss] at (2*\colS,11*-\rowH) {};
\node[hit]  at (3*\colS,11*-\rowH) {}; \node[num] at (3*\colS,11*-\rowH) {13$\times$};
\node[miss] at (4*\colS,11*-\rowH) {};
\node[rowlbl] at (0,12*-\rowH) {TSFM Opt. (G27)};
\node[miss] at (1*\colS,12*-\rowH) {};
\node[hit]  at (2*\colS,12*-\rowH) {}; \node[num] at (2*\colS,12*-\rowH) {3.3$\times$};
\node[miss] at (3*\colS,12*-\rowH) {};
\node[miss] at (4*\colS,12*-\rowH) {};
\end{tikzpicture}
\caption{Evidence heatmap: fourteen extension studies (rows) $\times$ four evaluation dimensions (cols)}
\label{tab:dimensions}
\end{figure}

\paragraph{Additional cases.} Six further cases (multi-turn artifact reuse, weight-internalized tool knowledge, confidence-gated routing, asset-class extensions to batteries and transformers, caching trustworthiness, and a multi-modal visual-inspection extension) are presented in Appendix~\ref{appx:more-cases}; each follows the same evidence-and-lesson structure as the cases above.

\section{Implications for Leaderboard Design}

The position translates into three concrete proposals for how
leaderboards should be structured.

\paragraph{Proposal 1: Declared configuration columns.} Beyond Model and
Pass$^{1}$, submissions should declare: Architecture
(Plan-Execute, Supervisor-Specialist, Solo, etc.), Reasoning Mode
(off, on, adaptive), Retrieval Strategy (none, single-pass,
multi-hop, internalized), Prompt-Constraint Level (none, light,
constrained), and Verifier Type (none, conclusion-only,
trajectory-only, both). Each is a non-empty axis whose value affects how
performance should be attributed; conflating across them produces
misleading rankings. The SmartGridBench study
\citep{bhandari2025smartgridbench} provides a direct empirical motivation
for this proposal: in a $2{,}420$-trajectory experiment over the
plan-execute base agent, varying transport (direct vs.\ MCP) and
orchestration (Plan-Execute vs.\ Verified PE vs.\ Self-Ask)
independently shows that MCP standardization adds latency without
quality gain, while orchestration changes alone raise pass rate from
$43.2\%$ to $55.5\%$. A leaderboard collapsing across these two axes
attributes orchestration wins to transport, or vice versa.

\paragraph{Proposal 2: Layered presentation.} A leaderboard reader
should see (Layer 1) a small headline table with PV rank,
(Layer 2) a cost-Pareto plot showing where each configuration
sits, (Layer 3) drill-down panels per tier, and (Layer 4)
significance and confidence intervals. Each layer answers a different
question. Headline tables should not aggregate beyond their resolution;
details should not bury the headline.

\paragraph{Proposal 3: Required submission elements.} Multi-run
variance, hardware disclosure, declared tier coverage, raw trajectories.
The field also needs two community artifacts: a shared rule pipeline
for judge-independent verification, and an adversarial-perturbation
suite for Criterion C.

\label{sec:agenda}

\paragraph{Alternative views.}
\textit{(i) Aggregate scores are sufficient for the model-comparison
purpose leaderboards actually serve.} On a per-model basis, an
aggregate score is a useful summary; our claim is narrower, that
\textit{rankings} derived from such scores do not transfer.
\textit{(ii) Predictive validity is itself measured by a single
correlation and so reproduces the problem.} It does not: predictive
validity is reported across three OOD criteria with explicit
thresholds and bootstrap CIs (Section~\ref{sec:falsification}), so a
single number cannot hide the variance behind it.
\textit{(iii) HELM and Dynabench already address multi-dimensionality.}
They address it for single-shot models; agents introduce trajectory-
and orchestration-level axes that those frameworks do not score
(Figure~\ref{fig:twelve-tiers}).

\section{Summary and Outlook}

This paper consolidated the largest coordinated extension of one
agent benchmark we are aware of: fourteen parallel implementation
studies, roughly six thousand judged trajectories, six extension axes
(asset class, orchestration, knowledge/retrieval, infrastructure,
reasoning mode, evaluation methodology) and twelve evaluation tiers.
The consolidated picture is that aggregate-score leaderboards
systematically underspecify the dimensions on which deployed LLM
agents are judged; the corrective is to rank by predictive validity
rather than in-sample mean, and to report on a multi-tier apparatus
that exposes the deployment-relevant dimensions current single-shot
multi-metric frameworks (HELM, Dynabench, behavioral testing) do not
score. The position is concrete enough to be falsified through three
OOD criteria with explicit thresholds.

Drawing the fourteen studies' forward-looking suggestions together
(Appendix~\ref{appx:forward}), four field-level recommendations
emerge. \textbf{1) Declare configurations, not just models.} Submission
fields for architecture, reasoning mode, retrieval strategy,
prompt-constraint level, and verifier type let readers attribute
performance to its cause. \textbf{2) Rank by transfer, not by mean.}
Report predictive-validity scores across at least one OOD criterion;
treat the in-sample mean as one column among many.
\textbf{3) Require a judge-independent anchor.} Every leaderboard
should have at least one trajectory-level deterministic verifier
(rule pipeline, DAG oracle, or similar) so judge drift is detectable.
\textbf{4) Adopt persistent, non-stdio infrastructure for the
benchmark itself.} Three of fourteen studies independently identified
MCP-stdio overhead as a dominant latency floor; a serious
deployment-oriented benchmark cannot leave protocol overhead
conflated with reasoning ability. These four moves do not require a
new benchmark; they require a different relationship between
leaderboards and the deployments they advise.

\section*{Limitations}

We are explicit about what this paper does not establish.

\paragraph{Empirical validation is future work.} The predictive-validity
claim has not been tested at scale by us. We have specified the
experiment but have not run it. The position is supported by
convergent architectural-sensitivity evidence across fourteen implementation studies, not by a
controlled randomized trial. Reviewers should evaluate the paper as a
synthesis and position, not as an empirical finding.

\paragraph{Domain specificity.} All evidence is drawn from industrial
asset operations and maintenance, specifically AssetOpsBench and its
extensions. Whether the twelve-tier apparatus generalizes to other
MCP-based domains (scientific assistants, customer service, code
generation) is open. We frame the apparatus as a hypothesis tested within
one domain.

\paragraph{Tier independence is asserted, not tested.} We claim the
twelve tiers are roughly orthogonal. We have not empirically tested this;
doing so requires the experimental work the position calls for.
Reviewers should read tier-orthogonality as a working hypothesis.

\paragraph{Industrial deployment validity gap.} All proposed
predictive-validity criteria remain internal to AssetOpsBench's
measurement apparatus. We do not yet have data linking framework
rankings to real deployed-system outcomes (operator override rate,
incident reduction, false-alarm rate). Closing this gap requires
industry partnerships beyond what this paper resolves.

\paragraph{Implementation-study epistemic status.} The fourteen
implementation studies are unpublished implementation reports, not
peer-reviewed publications. Their value here is convergence under
architectural diversity, not the independent peer review each would
warrant as a standalone empirical contribution.

\section*{Ethical Considerations}

This paper makes no use of human-subjects data, personally-identifiable
information, or sensitive deployed-system traces. The implementation
studies we synthesize use the publicly released AssetOpsBench dataset
(Apache-2.0 license) and synthetic or publicly-available industrial
data sources (e.g., NASA PCOE Li-ion cycling data). The position the
paper advances; that leaderboards should measure predictive
validity; has a potential downside: such measurements are more
expensive than aggregate scoring and could concentrate evaluation
capacity in well-resourced institutions. We flag this as a
field-organization concern and recommend that reference
adversarial-perturbation suites and reference rule pipelines be
maintained as community artifacts to mitigate the gap.

\bibliography{references}

@article{patel2025assetopsbench,
  title={Assetopsbench: Benchmarking ai agents for task automation in industrial asset operations and maintenance},
  author={Patel, Dhaval and Lin, Shuxin and Rayfield, James and Zhou, Nianjun and Shyalika, Chathurangi and Yarrabothula, Suryanarayana R and Vaculin, Roman and Martinez, Natalia and O'donncha, Fearghal and Kalagnanam, Jayant},
  journal={arXiv preprint arXiv:2506.03828},
  year={2025}
}

@inproceedings{
wang2026mcpbench,
title={{MCP}-Bench: Benchmarking Tool-Using {LLM} Agents with Complex Real-World Tasks via {MCP} Servers},
author={Zhenting Wang and Qi Chang and Hemani Patel and Shashank Biju and Cheng-En Wu and Quan Liu and Aolin Ding and Alireza Rezazadeh and Ankit Shah and Yujia Bao and Eugene Siow},
booktitle={The Fourteenth International Conference on Learning Representations},
year={2026},
url={https://openreview.net/forum?id=fe8mzHwMxN}
}

@article{luo2025mcpuniverse,
  title={Mcp-universe: Benchmarking large language models with real-world model context protocol servers},
  author={Luo, Ziyang and Shen, Zhiqi and Yang, Wenzhuo and Zhao, Zirui and Jwalapuram, Prathyusha and Saha, Amrita and Sahoo, Doyen and Savarese, Silvio and Xiong, Caiming and Li, Junnan},
  journal={arXiv preprint arXiv:2508.14704},
  year={2025}
}

@article{meta2025are,
  title={Are: Scaling up agent environments and evaluations},
  author={Froger, Romain and Andrews, Pierre and Bettini, Matteo and Budhiraja, Amar and Cabral, Ricardo Silveira and Do, Virginie and Garreau, Emilien and Gaya, Jean-Baptiste and Lauren{\c{c}}on, Hugo and Lecanu, Maxime and others},
  journal={arXiv preprint arXiv:2509.17158},
  year={2025}
}

@article{shen2024taskbench,
  title={Taskbench: Benchmarking large language models for task automation},
  author={Shen, Yongliang and Song, Kaitao and Tan, Xu and Zhang, Wenqi and Ren, Kan and Yuan, Siyu and Lu, Weiming and Li, Dongsheng and Zhuang, Yueting},
  journal={Advances in Neural Information Processing Systems},
  volume={37},
  pages={4540--4574},
  year={2024}
}

@article{bandel2026exgentic,
  title={General agent evaluation},
  author={Bandel, Elron and Yehudai, Asaf and Eden, Lilach and Sagron, Yehoshua and Perlitz, Yotam and Venezian, Elad and Razinkov, Natalia and Ergas, Natan and Ifergan, Shlomit Shachor and Shlomov, Segev and others},
  journal={arXiv preprint arXiv:2602.22953},
  year={2026}
}

@article{odonncha2026condition,
  title={Evidence-Driven Reasoning for Industrial Maintenance Using Heterogeneous Data},
  author={O'Donncha, Fearghal and Zhou, Nianjun and Martinez, Natalia and Rayfield, James T and Heath III, Fenno F and Langbridge, Abigail and Vaculin, Roman},
  journal={arXiv preprint arXiv:2603.08171},
  year={2026}
}

@inproceedings{jimenez2023swebench,
  title={Swe-bench: Can language models resolve real-world github issues?},
  author={Jimenez, Carlos E and Yang, John and Wettig, Alexander and Yao, Shunyu and Pei, Kexin and Press, Ofir and Narasimhan, Karthik},
  booktitle={International Conference on Learning Representations},
  volume={2024},
  pages={54107--54157},
  year={2024}
}

@article{yao2024taubench,
  title={tau-bench: A Benchmark for Tool-Agent-User Interaction in Real-World Domains},
  author={Yao, Shunyu and Shinn, Noah and Razavi, Pedram and Narasimhan, Karthik},
  journal={arXiv preprint arXiv:2406.12045},
  year={2024}
}

@article{zheng2023mtbench,
  title={Judging llm-as-a-judge with mt-bench and chatbot arena},
  author={Zheng, Lianmin and Chiang, Wei-Lin and Sheng, Ying and Zhuang, Siyuan and Wu, Zhanghao and Zhuang, Yonghao and Lin, Zi and Li, Zhuohan and Li, Dacheng and Xing, Eric and others},
  journal={Advances in neural information processing systems},
  volume={36},
  pages={46595--46623},
  year={2023}
}

@misc{gowda2025agentopsbench,
  title  = {AgentOpsBench: High-Throughput Agentic AI for Battery Analytics},
  author = {Gowda, Siddharth and Bhatt, Rushin and Agrawal, Aryaman and Li, Winston},
  year   = {2026},
  note   = {HPML Spring 2026 Final Project, Columbia University},
  url    = {https://github.com/siddharthgowda/AssetOpsBench}
}

@techreport{li2025multiturn,
  title={Towards Multi-Turn Dialog Systems for Industrial Asset Operations and Maintenance},
  author={Li, Rujing and Bai, Yitong and Li, Chengrui and Li, Rui},
  year={2026},
  note   = {HPML Spring 2026 Final Project, Columbia University},
  url = {https://github.com/Coderlicr/Multi-Turn-AssetOps}
}

@misc{li2025profiling,
  title  = {Profiling and Optimizing the AssetOpsBench Plan-Execute Pipeline},
  author = {Li, Shen and Xu, Charles and Li, Ann and Cahill, Caroline},
  year   = {2026},
  note   = {HPML Spring 2026 Final Project, Columbia University},
  url    = {https://github.com/jasonlee-1024/AssetOpsBench/tree/main}
}

@techreport{li2025skillsknowledge,
  title={Skills and Knowledge Plugin {MCP} Servers for Optimized Industrial {O\&M} Agents},
  author={Li, Andrew and Natarajan, Kirthana and On, Thai and Maturi, Trisha and Bhuvanesh, Yeshitha},
  year={2026},
  note = {HPML Spring 2026 Final Project, Columbia University},
  url = {https://github.com/kmn01/AssetOpsBench/tree/dev}
}

@techreport{mazeeva2025skillknowledge,
  title={Skill-Knowledge-Augmented Agents on {AssetOpsBench}},
  author={Mazeeva, Vera and Abbaszadeh, Mana and Arora, Shrey and Shejwal, Sanskruti},
  institution={Technical Report},
  year={2026},
  url = {https://github.com/shreyarora2198/AssetOpsBench/tree/team14-final}
}

@article{merchant2025temporal,
  title={Evaluating Temporal Semantic Caching and Workflow Optimization in Agentic Plan-Execute Pipelines},
  author={Merchant, Alimurtaza Mustafa and Veera, Krish and Goyla, Sajal Kumar and Bhure, Shambhawi and Patel, Dhaval and Maghraoui, Kaoutar El},
  journal={arXiv preprint arXiv:2605.20630},
  year={2026}
}

@misc{group202025qlora,
title = {Internalizing MCP Tool Knowledge in Small LLMs via QLoRA Fine-Tuning
},
author = {Agarwal, Tanmay and Shemla, Yuval and Yakobe, Ayal},
year = {2026},
note = {HPML Spring 2026 Final Project, Columbia University},
url = {https://github.com/YuvalShemla/hpml-2026-project.git}
}

@techreport{kumar2025transformer,
  title={Extending {AssetOpsBench}: Smart Grid Transformer Integration and Automated Scenario Generation},
  author={Kumar, Sagar Chethan and Kanathur, Rohith and Kapoor, Ananya and Bahl, Dev},
  institution={Technical Report},
  year={2026},
  note = {HPML Spring 2026 Final Project, Columbia University},
  url = {https://github.com/Rohith-Kanathur/AssetOpsBench}
}

@article{li2025phmforge,
  title={PHMForge: Evaluating LLM Agents on Industrial Prognostics through MCP-Native, Algorithm-Grounded Tools},
  author={Feng, Tianjun and Chen, Yunfeng and Tsai, Chun-Yi and Sun, Yihan and Das, Ayan and Maghraoui, Kaoutar El and Lin, Shuxin and Patel, Dhaval},
  journal={arXiv preprint arXiv:2604.01532},
  year={2026}
}

@techreport{bhandari2025smartgridbench,
  title={{SmartGridBench}: {MCP}-Based Industrial Agent Benchmarking for Smart Grid Transformer Operations},
  author={Bhandari, Akshat and Fan, Aaron and Rathod, Tanisha and Xin, Wei Alexander},
  institution={Technical Report},
  year={2026}
}

@techreport{jebbouri2025fmsrbottleneck,
  title={Performance Optimization for {FMSR} Call-Matrix Bottleneck and Quantization Substitution in Industrial Agentic Benchmarks},
  author={Jebbouri, Yassine and Maes, Darief Rida and Rachakonda, Shriya Aishani and Iyer, Vivek G.},
  institution={Technical Report},
  year={2026}
}

@techreport{go2025tsfmprofile,
  title={Profiling and Optimizing the {TSFM} {MCP} Server},
  author={Go, Sally and Colman, Sam and Kwon, Byeolah and Pasiecznik, Tomas},
  institution={Technical Report},
  year={2026}
}

@techreport{vinod2025tsfmoptim,
  title={Performance Optimization of the {TSFM} Agent in an Industrial Agentic Benchmark},
  author={Vinod, Alisha and Ajai, Thomas and Ang, Jonathan and Vijayakumar, Sanjaii},
  institution={Technical Report},
  year={2026}
}

@techreport{sheikh2025vi,
  title={Multi-Modal Agent Inference Optimization for Industrial Asset Operations: Extending {AssetOpsBench} with a Visual Inspection Agent via {MCP}},
  author={Sheikh, Amaan and Upganlawar, Aman and Rajkondawar, Madhav and Chen, Yang-Jung},
  institution={Technical Report},
  year={2026}
}

@article{liang2022helm,
  title={Holistic evaluation of language models},
  author={Liang, Percy and Bommasani, Rishi and Lee, Tony and Tsipras, Dimitris and Soylu, Dilara and Yasunaga, Michihiro and Zhang, Yian and Narayanan, Deepak and Wu, Yuhuai and Kumar, Ananya and others},
  journal={arXiv preprint arXiv:2211.09110},
  year={2022}
}

@inproceedings{kiela2021dynabench,
  title={Dynabench: Rethinking benchmarking in NLP},
  author={Kiela, Douwe and Bartolo, Max and Nie, Yixin and Kaushik, Divyansh and Geiger, Atticus and Wu, Zhengxuan and Vidgen, Bertie and Prasad, Grusha and Singh, Amanpreet and Ringshia, Pratik and others},
  booktitle={Proceedings of the 2021 conference of the North American chapter of the Association for Computational Linguistics: human language technologies},
  pages={4110--4124},
  year={2021}
}

@inproceedings{ribeiro2020checklist,
  title={Beyond accuracy: Behavioral testing of NLP models with CheckList},
  author={Ribeiro, Marco Tulio and Wu, Tongshuang and Guestrin, Carlos and Singh, Sameer},
  booktitle={Proceedings of the 58th annual meeting of the association for computational linguistics},
  pages={4902--4912},
  year={2020}
}

@inproceedings{bowman2021will,
  title={What will it take to fix benchmarking in natural language understanding?},
  author={Bowman, Samuel and Dahl, George},
  booktitle={Proceedings of the 2021 Conference of the North American Chapter of the Association for Computational Linguistics: Human Language Technologies},
  pages={4843--4855},
  year={2021}
}

@article{raji2021everything,
  title={AI and the everything in the whole wide world benchmark},
  author={Raji, Inioluwa Deborah and Bender, Emily M and Paullada, Amandalynne and Denton, Emily and Hanna, Alex},
  journal={arXiv preprint arXiv:2111.15366},
  year={2021}
}

@inproceedings{ethayarajh2020utility,
  title={Utility is in the eye of the user: A critique of NLP leaderboards},
  author={Ethayarajh, Kawin and Jurafsky, Dan},
  booktitle={Proceedings of the 2020 Conference on Empirical Methods in Natural Language Processing (EMNLP)},
  pages={4846--4853},
  year={2020}
}

@inproceedings{recht2019imagenet,
  title={Do imagenet classifiers generalize to imagenet?},
  author={Recht, Benjamin and Roelofs, Rebecca and Schmidt, Ludwig and Shankar, Vaishaal},
  booktitle={International conference on machine learning},
  pages={5389--5400},
  year={2019},
  organization={PMLR}
}

@article{dehghani2021benchmark,
  title={The benchmark lottery},
  author={Dehghani, Mostafa and Tay, Yi and Gritsenko, Alexey A and Zhao, Zhe and Houlsby, Neil and Diaz, Fernando and Metzler, Donald and Vinyals, Oriol},
  journal={arXiv preprint arXiv:2107.07002},
  year={2021}
}

@article{xuan2026interactive,
  title={Interactive Evaluation Requires a Design Science},
  author={Xuan, Keyang and Song, Peiyang and Lu, Pan and Han, Pengrui and Li, Wenkai and Zhang, Zhenyu and He, Zexue and Hua, Wenyue and Li, Manling and You, Jiaxuan and others},
  journal={arXiv preprint arXiv:2605.17829},
  year={2026}
}

@article{patel2026codsretro,
  title={Results and Retrospective Analysis of the CODS 2025 AssetOpsBench Challenge},
  author={Patel, Dhaval and Shyalika, Chathurangi and Yarrabothula, Suryanarayana Reddy and Yue, Ling and Lin, Shuxin and Zhou, Nianjun and Rayfield, James},
  journal={arXiv preprint arXiv:2605.08518},
  year={2026}
}

@article{kalaitzidis2026evaltrap,
  title={The Evaluation Trap: Benchmark Design as Theoretical Commitment},
  author={Kalaitzidis, Theodore J},
  journal={arXiv preprint arXiv:2605.14167},
  year={2026}
}

\clearpage
\appendix

\section{Additional Convergent-Sensitivity Cases}
\label{appx:more-cases}

The following five cases complement the four in Section~\ref{sec:triangulation}.
Each follows the same evidence-and-lesson structure as the body cases.

\paragraph{Orchestration: multi-turn artifact reuse.}

A multi-turn study replaced AssetOpsBench's single-agent plan-execute
baseline with a Supervisor-Specialist (SS) architecture that reuses
tool outputs across turns, plus an SSA variant that adds parallel-tool
execution \citep{li2025multiturn}. SS dropped tool-time share of
wall-clock from $47.3\%$ to $26.3\%$, planning effectiveness rose from
$0.559$ to $0.791$, schema-failure rate fell $68.7\%$, and turns 2--5
ran $4.2\times$ faster than turn 1 (34.3\,s vs.\ 145.4\,s mean). The
parallel-tool variant, however, inflated token consumption (3.62M
vs.\ 3.32M for sequential SS, both above the 2.55M plan-execute
baseline) and worsened tail latency. Cross-turn artifact-reuse rate is
a deployment-cost metric invisible to single-turn benchmarks, and the
SSA counter-finding shows that a win on one axis can be a loss on
another.

\paragraph{Reasoning mode: weight-internalized tool knowledge.}

A QLoRA fine-tuning study internalized AssetOpsBench's tool catalogs
into the adapter weights of a Gemma-4-E4B model, eliminating the
${\sim}2{,}400$-token schema that the informed baseline prepends to
every prompt \citep{group202025qlora}. The fine-tuned model reached
AT-F1 $0.65$ versus the baseline's $0.47$ and judge score $3.88$
versus $2.88$, with an $82.6\%$ reduction in input tokens. A
catastrophic-forgetting probe, however, exposed a longitudinal cost:
Gemma retained $79.8\%$ of pre-training MCQ accuracy, while Qwen3-4B
under the same protocol retained only $61.3\%$. The implication
generalizes beyond this study: any weight-modifying optimization can
shift behavior on tasks the benchmark does not test, so a retention
sub-metric belongs in the failure-mode panel whenever a submitted
configuration involves fine-tuning.

\paragraph{Reasoning mode: confidence-gated routing.}

A skill-knowledge-augmented agent study audited baseline AssetOpsBench
trajectories and found that $35.5\%$ contained unnecessary calls to
the expensive Deep-TSFM tool, $96\%$ of them on wrong-domain
scenarios \citep{mazeeva2025skillknowledge}. The proposed fix routes
around Deep-TSFM when an FMSR confidence threshold $\theta$ is met; a
sweep over $\theta \in \{0.5,\dots,0.95\}$ identified $\theta=0.8$ as
the operating point. At that threshold, overall-correct rose from
$13.0\%$ to $30.4\%$, hallucination fell from $93.5\%$ to $35.6\%$,
and agent-sequence correctness rose from $6.5\%$ to $88.9\%$. The same
intervention shows up as a tool-hygiene improvement, an
adaptive-routing property, and a hallucination reduction at once; the
cross-cutting pattern is evidence that the measurement dimensions
interact whenever a configuration is optimized for deployment.

\paragraph{Asset class: batteries and transformers.}

Two extensions tested whether the AssetOpsBench framework actually
generalizes beyond its HVAC origin. One added a Smart-Grid Transformer
asset with four IEC-standards-grounded tools and an automated
scenario-generator \citep{kumar2025transformer}; the other added a
Battery-analytics MCP server with eleven tools seeded by NASA PCoE
cycling data \citep{gowda2025agentopsbench}. Both reported $6\times$
to $8\times$ end-to-end speedups under profiling-driven optimization,
with quality preserved in the transformer case (composite score
$74.2 \pm 1.9$ versus baseline $73.8 \pm 3.0$). The battery study
passed 8 of 11 scenarios at $7.4$\,s, against $44.6$\,s on the
10-cell fleet query baseline. ``Does the framework extend?'' is a
deployment question that no in-domain HVAC leaderboard can answer;
cross-domain transfer must be a reportable axis.

\paragraph{Knowledge augmentation: caching trustworthiness.}

A temporal-semantic caching study layered a parameter-aware cache
over the AssetOpsBench plan-execute pipeline, with a Volatile / Static
/ Relative / Anchored temporal classifier and DAG-layered parallel
execution \citep{merchant2025temporal}. The combined system achieved
$3.48\times$ end-to-end speedup over 80 paraphrase-tier queries, with
a median $30.6\times$ on cache hits. The cache's hit-decision F1
ceiling was $0.64$ (recall $0.5625$, precision $0.75$), limited by
parameter-collision false positives in industrial queries where two
syntactically similar requests target different physical assets.
``Caching helps'' and ``caching is safe'' are different claims that
require different metrics; cache-induced hallucination must be
tracked, not assumed away.

\paragraph{Asset class (multi-modal): visual inspection extension.}

A visual inspection extension added a Vision-Language MCP server to
AssetOpsBench, covering 22 scenarios across four asset classes (pumps,
power transformers, induction motors, wind-turbine blades) drawn from
public image datasets that AssetOpsBench's original text+timeseries
benchmark could not address \citep{sheikh2025vi}. The study sweeps ten
serving variants on a single L4 GPU for Qwen2.5-VL-7B and
Llama-3-LLaVA-NeXT-8B, isolating the quantization-to-serving stack.
Three findings cut against common practice. First, AWQ W4A16 with
domain-specific calibration was the only variant that improved both
speed and judge accuracy on the primary model (Qwen pass rate
$0.48 \to 0.82$ with $1.99\times$ latency reduction; the same
configuration with generic ultrachat calibration plateaued at $0.73$).
Second, an aggressive vLLM serving bundle combining FP8 KV-cache,
chunked prefill, and prefix caching collapsed Qwen's output entirely
($0/44$, response collapse) while passing on Llama; FP8 KV-cache
corruption of long visual-token prefixes is the suspected cause.
Third, downscaling input images to 512 pixels produced latency
regressions on both model families, indicating the resolution knee
for fixed-tile vision encoders sits below 512 px. The
broader lesson generalizes: in multi-modal agentic deployments,
quantization and serving choices interact with modality-specific
encoder behavior in ways that standard NLP benchmarks cannot detect.

\section{Tier Definitions}
\label{appx:tier-defs}

Each tier below names a measurement dimension, the source benchmark(s)
that surface it as a first-class metric, and a representative metric.

\begin{itemize}\setlength\itemsep{2pt}
  \item \textbf{T1 Success.} Headline pass/fail: AssetOpsBench's Pass$^{1}$ and Pass$^{k}$, ARE/Gaia2's DAG-Pass, the six rubric dimensions $y_1$--$y_6$. The minimum reportable floor, uninformative beyond it.
  \item \textbf{T2 Tool-Call Hygiene.} MCP-Bench's tool-name validity, schema compliance, execution success, dependency-order correctness. Non-redundant because hygiene failures occur on otherwise-successful trajectories.
  \item \textbf{T3 Planning Quality.} MCP-Bench's five LLM-judged planning axes plus TaskBench's ROUGE on decomposition, Node/Edge F1, chain-order NED \citep{shen2024taskbench}. Two configurations can succeed by different planning processes.
  \item \textbf{T4 Capability Axes.} ARE/Gaia2's seven axes (execution, search, adaptability, time, ambiguity, agent-to-agent, noise) plus TaskBench's Node/Chain/DAG stratification. Capability profile differs across similarly-ranked models.
  \item \textbf{T5 Cost \& Efficiency.} MCP-Universe and Gaia2's \$/scenario, step count, latency, budget-scaling curves. Cost-quality Pareto position determines deployment viability.
  \item \textbf{T6 Failure Modes.} AssetOpsBench's 14 MAST failure modes augmented with emergent clusters, distractor robustness, recovery rate. Failure modes inform mitigation design beyond pass-rate signal.
  \item \textbf{T7 Integrity \& Reproducibility.} Multi-run variance, prompt-shuffle averaging (MCP-Bench), validation/test split, judge--human inter-rater agreement.
  \item \textbf{T8 Deployment Infrastructure.} Latency decomposition, MCP-stdio overhead, subprocess-spawn count, optimization-impact ablation, cross-domain transfer. Surfaced by the battery-domain extension \citep{gowda2025agentopsbench}.
  \item \textbf{T9 Multi-Turn Dialog.} Cross-turn artifact reuse, per-turn cost dynamics, context-bloat trade-offs, multi-turn reliability, dialog-level Pass$^{1}$. Surfaced by \citep{li2025multiturn}.
  \item \textbf{T10 Reasoning Mode.} Per-phase reasoning cost attribution, per-rubric-dimension reasoning sensitivity, adaptive routing precision/recall, reasoning-quality-adjusted latency. Surfaced by \citep{li2025profiling}.
  \item \textbf{T11 Knowledge Augmentation.} Retrieval recall, multi-hop depth, embedding-index quality, skill-marketplace coverage and selection accuracy. Surfaced by \citep{li2025skillsknowledge}.
  \item \textbf{T12 Evidence Grounding \& Verification.} Judge-independent governance (Condition Agreement Rate, ARE hard/causality/timing violations), Unsupported Claim Rate, reasoning stability, verifier-type declarations. The most novel tier, drawn from systems that move beyond LLM-judge-only evaluation \citep{odonncha2026condition,meta2025are}.
\end{itemize}

\section{Empirical Evidence of Rank Instability}
\label{appx:rankevidence}

\begin{figure*}[t]
\centering
\definecolor{rhoblue}{HTML}{2E75B6}
\definecolor{rhored}{HTML}{C0392B}
\definecolor{rhogreen}{HTML}{1E8449}
\definecolor{thrline}{HTML}{C00000}
\begin{tikzpicture}[
  rowlabel/.style={font=\scriptsize, anchor=east, inner sep=2pt},
  tick/.style={font=\tiny, text=gray!55!black, anchor=north, inner sep=1pt},
  axlabel/.style={font=\scriptsize\bfseries, anchor=north},
  point/.style={circle, fill=#1, draw=#1!50!black, line width=0.2mm, minimum size=4.5pt, inner sep=0},
  cibar/.style={line width=0.45mm, #1},
  annot/.style={font=\tiny\itshape, text=gray!55!black, anchor=west, inner sep=2pt}
]
\def\scl{3.5}
\draw[gray!50, line width=0.25mm] (-1*\scl, 0.3) -- (1*\scl, 0.3);
\foreach \v in {-1,-0.5,0,0.5,1}{
  \draw[gray!50, line width=0.18mm] (\v*\scl, 0.3) -- (\v*\scl, 0.42);
  \node[tick] at (\v*\scl, 0.28) {$\v$};
}
\foreach \v in {-1,-0.5,0,0.5,0.85,1}{
  \draw[gray!25, dashed, line width=0.15mm] (\v*\scl, 0.3) -- (\v*\scl, -3.6);
}
\draw[thrline, dashed, line width=0.5mm] (0.85*\scl, 0.4) -- (0.85*\scl, -3.6);
\node[font=\tiny, text=thrline, anchor=south, rotate=90] at (0.85*\scl - 0.06, -1.6) {$\rho=0.85$ falsification threshold};

\def\rH{0.62}
\node[rowlabel] at (-1*\scl - 0.05, 0*-\rH - 0.4) {CODS planning, $n{=}20$};
\draw[cibar=rhoblue] (0.35*\scl, 0*-\rH - 0.4) -- (0.87*\scl, 0*-\rH - 0.4);
\draw[cibar=rhoblue] (0.35*\scl, 0*-\rH - 0.5) -- (0.35*\scl, 0*-\rH - 0.3);
\draw[cibar=rhoblue] (0.87*\scl, 0*-\rH - 0.5) -- (0.87*\scl, 0*-\rH - 0.3);
\node[point=rhoblue] at (0.69*\scl, 0*-\rH - 0.4) {};
\node[annot] at (1*\scl + 0.1, 0*-\rH - 0.4) {$\rho{=}0.69$};

\node[rowlabel] at (-1*\scl - 0.05, 1*-\rH - 0.4) {CODS execution, $n{=}13$};
\draw[cibar=rhored] (-0.64*\scl, 1*-\rH - 0.4) -- (0.45*\scl, 1*-\rH - 0.4);
\draw[cibar=rhored] (-0.64*\scl, 1*-\rH - 0.5) -- (-0.64*\scl, 1*-\rH - 0.3);
\draw[cibar=rhored] (0.45*\scl, 1*-\rH - 0.5) -- (0.45*\scl, 1*-\rH - 0.3);
\node[point=rhored] at (-0.13*\scl, 1*-\rH - 0.4) {};
\node[annot] at (1*\scl + 0.1, 1*-\rH - 0.4) {$\rho{=}{-}0.13$};

\node[rowlabel] at (-1*\scl - 0.05, 2*-\rH - 0.4) {Exgentic cross-benchmark range};
\draw[cibar=rhogreen] (0.32*\scl, 2*-\rH - 0.4) -- (0.85*\scl, 2*-\rH - 0.4);
\draw[cibar=rhogreen] (0.32*\scl, 2*-\rH - 0.5) -- (0.32*\scl, 2*-\rH - 0.3);
\draw[cibar=rhogreen] (0.85*\scl, 2*-\rH - 0.5) -- (0.85*\scl, 2*-\rH - 0.3);
\node[annot] at (1*\scl + 0.1, 2*-\rH - 0.4) {6 bench.};

\node[rowlabel] at (-1*\scl - 0.05, 3*-\rH - 0.4) {Proposed pilot ($n{=}80$, target)};
\draw[gray!50, line width=0.3mm] (0.55*\scl, 3*-\rH - 0.4) -- (0.95*\scl, 3*-\rH - 0.4);
\node[point=gray, fill=white!90!black] at (0.75*\scl, 3*-\rH - 0.4) {};
\node[annot] at (1*\scl + 0.1, 3*-\rH - 0.4) {planned};

\node[axlabel] at (0, -3.3) {Public$\to$Hidden Spearman rank correlation};
\end{tikzpicture}
\caption{Public-to-hidden rank correlations with $95\%$ Fisher-$z$ CIs. Dashed red line is the $\rho=0.85$ falsification threshold (Section~\ref{sec:falsification}).}
\label{fig:rho-forest}
\end{figure*}
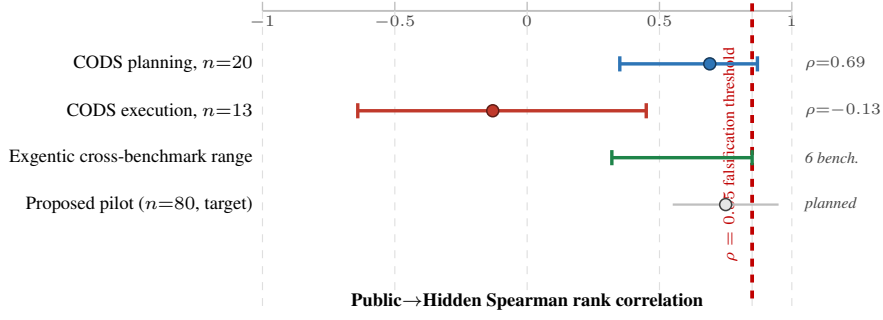

Figure~\ref{fig:rho-forest} aggregates the published rank-correlation
evidence cited above. We plot each correlation as a point estimate
with a $95\%$ confidence interval (Fisher $z$-transform) and a vertical
reference line at $\rho=0.85$; the falsification threshold from
Section~\ref{sec:falsification}. Three observations follow. First, the
execution-track correlation from \citet{patel2026codsretro} is
statistically indistinguishable from zero: with $n=13$, the $95\%$ CI
spans roughly $[-0.64,+0.45]$, wide enough that the point estimate
$\rho=-0.13$ is informative only as ``not robustly positive.'' Second,
the planning-track correlation is robustly positive
($\rho=0.69, n=20$, CI $\approx[0.35,0.87]$), but its upper bound still
falls at the falsification threshold. Third, Exgentic's reported
cross-benchmark range \citep{bandel2026exgentic} of $0.32$--$0.85$
across six heterogeneous benchmarks straddles the threshold from below.
Taken together, the available empirical evidence is consistent with the
position; that public-to-hidden rank stability is not reliably above
$0.85$; without being strong enough to confirm it. The empirical study
in Section~\ref{sec:agenda} is designed to close this gap.

\section{Per-Study Extended Details}
\label{appx:per-study}

The fourteen implementation studies are documented in two stacked
tables. Table~\ref{tab:per-study-setup} reports the baseline
configuration each study compared against and the magnitude of
improvement obtained. Table~\ref{tab:per-study-interp} reports the
conclusion drawn from each study and its single most prominent
forward-looking suggestion. Group identifiers (G3--G30) follow the
labels in Figure~\ref{fig:extension-map}.

\begin{table*}[t]
\centering\scriptsize
\setlength{\tabcolsep}{5pt}
\renewcommand{\arraystretch}{1.45}
\arrayrulecolor{gray!40}
\begin{tabular}{>{\bfseries}p{1.55cm}p{2.95cm}p{4.4cm}p{5.2cm}}
\arrayrulecolor{tblHdrBg}\specialrule{0.7mm}{0pt}{0pt}
\rowcolor{tblHdrBg}
\textcolor{tblHdrFg}{Group} &
\textcolor{tblHdrFg}{Axis} &
\textcolor{tblHdrFg}{Baseline} &
\textcolor{tblHdrFg}{Magnitude vs.\ baseline} \\
\arrayrulecolor{tblHdrBg}\specialrule{0.5mm}{0pt}{0pt}
\arrayrulecolor{gray!40}
G3 Skills+KP & \axispill{axKnowT}\,Knowledge/Retrieval & Single-pass RAG (top-$k$=3, Llama-4-Maverick-17B) & $\sim$90\% vs.\ 50--68\% acc.; $4.5\text{--}10\times$ tokens; $\sim$$7\times$ latency \\
\rowcolor{tblRowAlt}
G5 Multi-Turn & \axispill{axOrchT}\,Orchestration & Single-agent plan-execute & Planning $+41.6\%$; completion $+30.4\%$; tool-time $47.3\%\to 26.3\%$; turn 1 vs.\ 2--5 $4.2\times$ \\
G7 Battery & \axispill{axAssetT}\,Asset Class & Plain plan-execute on per-cell MCP, no cache or batching & $6.06\times$ end-to-end; disk cache 7.0\,s $\to$ 0.002\,s ($\sim$3500$\times$) \\
\rowcolor{tblRowAlt}
G8 Transformer & \axispill{axEvalT}\,Evaluation Methodology & Sequential scenario generator, no cache, Llama-3.3-70B & $8\times$ at $N{=}50$; quality $74.2{\pm}1.9$ vs.\ $73.8{\pm}3.0$ (preserved) \\
G9 Temporal Cache & \axispill{axKnowT}\,Knowledge/Retrieval & No-cache plan-execute, Qwen3 emb.\ + reranker & $30.6\times$ on cache hits; $3.48\times$ overall; F1 cap $\sim$0.64 \\
\rowcolor{tblRowAlt}
G12 SmartGridBench & \axispill{axEvalT}\,Evaluation Methodology & AT-I direct in-process tools (Llama-3.1-8B/vLLM) & V-PE+Self-Ask: 0.620 mean / 55.5\% pass vs.\ 0.474 / 43.2\% \\
G14 Conf.\ Gate & \axispill{axReasT}\,Reasoning Mode & Static plan-execute on four MCP servers & Overall $13\%\to 30.4\%$; hallu.\ $93.5\%\to 35.6\%$; seq.\ $6.5\%\to 88.9\%$ \\
\rowcolor{tblRowAlt}
G16 FMSR Dispatch & \axispill{axInfraT}\,Infrastructure & Sequential FMSR $N{\times}M$ w/ cloud-hosted Llama-3.3-70B & Hedged $36\times$ (559\,s$\to$15.5\,s); INT4 3B: 0.675 vs.\ 0.660; DB prefetch hurts Wind Turbines (acc.\ $0.60\to 0.43$) \\
G19 TSFM Backend & \axispill{axInfraT}\,Infrastructure & Unoptimized TSFM (TTM) w/ HF Trainer & Forecasting $-33.3\%$; Chronos $12.8\times$ faster forecasting but $81\times$ slower on fine-tuning \\
\rowcolor{tblRowAlt}
G20 QLoRA & \axispill{axReasT}\,Reasoning Mode & Unfine-tuned Gemma-4-E4B with full schemas inline & AT-F1 $+0.18$; tokens $-82.6\%$; Qwen3 retains 61.3\% MCQ vs.\ Gemma 79.8\% \\
G21 Profiling & \axispill{axReasT}\,Reasoning Mode & Gemma-4-26B reasoning-off via vLLM/A100 & Clarity $+31$pp; completion $+16$pp; latency $+21.5\%$; planning $+41.9\%$ \\
\rowcolor{tblRowAlt}
G27 TSFM Opt. & \axispill{axInfraT}\,Infrastructure & Untouched TSFM agent w/ HF Trainer, CPU & ttm\_forward $-69\%$; model load $-89\%$; bf16 regresses $3.3\times\to 2.6\times$ \\
G30 PHMForge & \axispill{axEvalT}\,Evaluation Methodology & Text-RAG over telemetry-equivalent evidence (Claude Opus 4.6) & MCP vs.\ RAG: 80.6\% vs.\ 48.6\% (McNemar $p{=}0.002$); RUL: 100\%$\to$20\% \\
\rowcolor{tblRowAlt}
G23 Visual Inspect & \axispill{axAssetT}\,Asset Class (multi-modal) & FP16 Qwen2.5-VL-7B baseline on 22 visual scenarios (4 asset classes), L4 GPU & AWQ W4A16 + domain calibration: pass 0.48$\to$0.82, latency $9.4\,\text{s}\to 4.8\,\text{s}$ ($1.99\times$); L2 full-bundle FP8-KV: 0/44 (response collapse); L3 image\_512: latency regression \\
\arrayrulecolor{tblHdrBg}\specialrule{0.5mm}{0pt}{0pt}
\arrayrulecolor{black}
\end{tabular}
\caption{Per-study setup: baseline configuration and magnitude of improvement. Axis pill matches Figure~\ref{fig:extension-map}.}
\label{tab:per-study-setup}
\end{table*}

\begin{table*}[t]
\centering\scriptsize
\setlength{\tabcolsep}{5pt}
\renewcommand{\arraystretch}{1.45}
\arrayrulecolor{gray!40}
\begin{tabular}{>{\bfseries}p{1.55cm}p{2.95cm}p{5.05cm}p{4.55cm}}
\arrayrulecolor{tblHdrBg}\specialrule{0.7mm}{0pt}{0pt}
\rowcolor{tblHdrBg}
\textcolor{tblHdrFg}{Group} &
\textcolor{tblHdrFg}{Axis} &
\textcolor{tblHdrFg}{Conclusion} &
\textcolor{tblHdrFg}{Forward-looking} \\
\arrayrulecolor{tblHdrBg}\specialrule{0.5mm}{0pt}{0pt}
\arrayrulecolor{gray!40}
G3 Skills+KP & \axispill{axKnowT}\,Knowledge/Retrieval & Knowledge plugins act as inference-time scaling: accuracy at cost. & Binary classifier to route RAG vs.\ Knowledge Plugin. \\
\rowcolor{tblRowAlt}
G5 Multi-Turn & \axispill{axOrchT}\,Orchestration & Supervisor-specialist + artifact reuse beats single-agent plan-execute. & Parallel specialists with evidence summarization. \\
G7 Battery & \axispill{axAssetT}\,Asset Class & MCP-stdio subprocess overhead dominates inference. & Persistent (non-stdio) MCP server mode. \\
\rowcolor{tblRowAlt}
G8 Transformer & \axispill{axEvalT}\,Evaluation Methodology & Standards-grounded (IEC) auto scenario generation scales benchmark authoring. & Cross-model judge; human-annotation replacement. \\
G9 Temporal Cache & \axispill{axKnowT}\,Knowledge/Retrieval & Pure semantic similarity is structurally insufficient for parameter-rich queries. & Parameter-aware (entity, sensor, time-window) caching. \\
\rowcolor{tblRowAlt}
G12 SmartGridBench & \axispill{axEvalT}\,Evaluation Methodology & Protocol, orchestration, and evidence-accounting are separate variables. & Extend hosted-70B coverage; judge remaining 25 scenarios. \\
G14 Conf.\ Gate & \axispill{axReasT}\,Reasoning Mode & The strongest agent is the one that knows when a tool is unnecessary. & Learned confidence calibrator; per-asset thresholds. \\
\rowcolor{tblRowAlt}
G16 FMSR Dispatch & \axispill{axInfraT}\,Infrastructure & FMSR $N{\times}M$ is the dominant bottleneck; cross-asset prefetch transfer is non-trivial. & Hedged dispatch + INT4 small models for binary relevancy. \\
G19 TSFM Backend & \axispill{axInfraT}\,Infrastructure & No universal TSFM backbone; wins on one workflow are losses on another. & Standardized TSAD preprocessing; pluggable backends. \\
\rowcolor{tblRowAlt}
G20 QLoRA & \axispill{axReasT}\,Reasoning Mode & Tool knowledge moves from prompt to weights, but with retention cost. & Continual learning for new tools; rank-aware retention. \\
G21 Profiling & \axispill{axReasT}\,Reasoning Mode & Reasoning pays its cost in planning, returns it in downstream quality. & Adaptive thinking router (DistilBERT, 66M). \\
\rowcolor{tblRowAlt}
G27 TSFM Opt. & \axispill{axInfraT}\,Infrastructure & Agentic-ML optimization $\ne$ model optimization; bf16 hurts MLP-Mixer. & Persistent MCP via socket reuse; sensor batching. \\
G30 PHMForge & \axispill{axEvalT}\,Evaluation Methodology & Algorithm-grounded MCP tools are a necessary substrate; orchestration dominates failures. & High-performance agent serving; community scenario expansion. \\
\rowcolor{tblRowAlt}
G23 Visual Inspect & \axispill{axAssetT}\,Asset Class (multi-modal) & Domain-calibrated AWQ quantization is the rare optimization that improves both speed and judge accuracy; aggressive serving bundles (FP8-KV) can silently collapse VLM output quality. & Isolate FP8-KV from other vLLM flags; ablate image-resolution between 256--384 px; extend to chillers and AHUs when public image data appears. \\
\arrayrulecolor{tblHdrBg}\specialrule{0.5mm}{0pt}{0pt}
\arrayrulecolor{black}
\end{tabular}
\caption{Per-study interpretation: the conclusion the source report draws and the single forward-looking suggestion it most directly emphasizes.}
\label{tab:per-study-interp}
\end{table*}

\section{Convergence Patterns}
\label{appx:convergence}

Independent teams, working from the same base benchmark with different
architectures, converged on the same measurement gaps. We list five
convergence patterns we identified across the fourteen studies. Each
pattern is referenced by name in the main body of the paper. The
patterns are not mutually exclusive; many groups fall in more than one.

\paragraph{Pattern A: Plan-execute bottleneck localization.}
Groups G5, G14, G16, and G27 independently localize the same bottleneck
to FMSR or TSFM. G5 measures TSFM tool-time as the dominant share of
wall-clock (47.3\% to 26.3\% with supervisor routing). G14 audits the
baseline and finds 35.5\% of trajectories make redundant TSFM calls
(96\% wrong-domain). G16 measures FMSR $N\times M$ as 559s reduced to
15.5s under hedged dispatch. G27 isolates \texttt{ttm\_forward} as 93\%
of single-call latency. Four teams localize one bottleneck. The base
architecture is under-designed, not near-optimal.

\paragraph{Pattern B: Prompt-versus-weight tool knowledge.}
Groups G3, G14, G20, and G21 all improve accuracy by adding ``more
thinking'' along distinct axes. G3 adds retrieval (KP). G14 adds
confidence-gated routing. G20 internalizes the tool catalog into
adapter weights, eliminating 82.6\% of input tokens. G21 turns on
extended-thinking mode. These four are tiling the same trade-off
frontier from different sides. Prompt-time versus weight-time tool
knowledge is a tunable axis, not a fixed design choice.

\paragraph{Pattern C: MCP transport overhead.}
Groups G7, G9, G12, and G27 independently measure MCP-stdio subprocess
overhead as a dominant per-call latency floor. G7 reports a 5--6s
residual per call. G9 measures discovery as 2--3s per query, eliminated
by disk cache. G12 finds direct in-process tools (AT-I) outperform MCP
(AT-M) on both pass and judge mean. G27 measures model-loading $233$ms
$\to$ $26$ms via pre-loading. The base benchmark conflates protocol
overhead with reasoning ability.

\paragraph{Pattern D: Caching trustworthiness.}
Groups G3, G9, G16, and G27 each add a cache. G9 alone surfaces the
parameter-collision F1 ceiling at 0.64 as a structural failure mode.
``Caching helps'' and ``caching is safe'' are different claims and
require different metrics.

\paragraph{Pattern E: Scenario authoring as a binding constraint.}
Groups G7, G8, G12, and G30 all confront the same problem: the original
141-scenario corpus is small. G7 settles for 15 hand-authored battery
scenarios. G12 freezes a 36-scenario paper-grade set. G8 automates
scenario generation. G30 scales SME-authored scenarios to 99 across
eight asset classes. Scenario-corpus size is the binding constraint on
extension work in the field.

\section{Forward-Looking Suggestion Clusters}
\label{appx:forward}

The fourteen forward-looking suggestions cluster into four families with
strong cross-group convergence. Each cluster aggregates suggestions
across multiple studies into a single field-level recommendation.

\paragraph{Cluster I: Adaptive and learned routing.}
Five groups (G3, G9, G14, G16, G21) converge on the same idea. Some
queries need the expensive path; most do not; a static pipeline cannot
tell them apart. G3 proposes a learned RAG-versus-KP classifier. G9
calls for parameter-aware caching. G14 calls for a learned confidence
calibrator. G16 proposes adaptive ceiling-start dispatch. G21 proposes
a DistilBERT reasoning-mode router. The aggregate signal: a meta-router
should be a first-class abstraction in the framework.

\paragraph{Cluster II: Persistent, non-stdio MCP servers.}
Three groups (G7, G9, G27) identify subprocess spawn and stdio transport
as the residual latency floor and call for a persistent server. G7
proposes ``persistent MCP server modes.'' G9 proposes cache persistence
via FAISS. G27 proposes ``a persistent MCP server through socket
reuse.''

\paragraph{Cluster III: Cross-model and cross-asset generalization.}
Four groups (G8, G14, G16, G19) call for testing on more than one model
or more than one asset class. G8 proposes GPT-4, Llama-4, Mistral Large
coverage. G14 proposes pumps, bearings, motors, AHUs. G16 proposes
asset domains beyond chillers. G19 proposes an interchangeable
model-interface for additional TSFM backbones.

\paragraph{Cluster IV: Scenario-corpus expansion.}
Four groups (G8, G12, G20, G30) call for moving past the 141-scenario
corpus. G8 provides the automated tooling. G12 has 25 unjudged
scenarios pending. G20 proposes continual-learning protocols for new
tools. G30 has a public living-benchmark expansion plan.

The aggregate of these four clusters is exactly the position-paper
thesis. AssetOpsBench is under-specified along the same four axes that
all fourteen groups improvise around: adaptive routing, persistent
infrastructure, cross-model/asset coverage, and scenario authoring.

\section{Predictive-Validity Score: Methodological Notes}
\label{appx:pv}

The predictive-validity score
$\text{PV}(c) = \alpha\bar{Y}_c - \beta\sigma_{Y_c,\text{OOD}} - \gamma\,\text{IQR}(Y_c)$
is a proposal, not a finalized scoring rule. This appendix records
methodological notes for the empirical study we propose in
Section~\ref{sec:agenda}.

\paragraph{Weight fitting.} The weights $\alpha,\beta,\gamma$ are fit on
Criterion-A holdouts to maximize Spearman correlation between
predictive-validity rank and Criterion-B/C ranks. We propose a constrained
optimization over the simplex $\alpha+\beta+\gamma=1$ with all
$\geq 0$, and recommend reporting the full Pareto frontier rather than
a single point estimate.

\paragraph{Statistical power.} For 80 paired ranks, Spearman $\rho=0.85$
versus $\rho=0.95$ is distinguishable at $\alpha=0.01$ with power
above 0.9 (Fisher-$z$ transform). For 1{,}200 perturbed observations
(120 scenarios $\times$ top-10 configurations), mean-metric differences
of 0.05 are detectable with power above 0.95. The proposed pilot is
under-powered for fine subgroup analyses and we will report subgroup
results as exploratory only.

\paragraph{Pre-registered pilot decision rule.} If pilot train--holdout
Spearman exceeds $0.95$ across all three architectures, we revise the
central claim downward: AssetOpsBench rankings appear highly stable at
i.i.d.\ holdout and the full study is unlikely to clear the
$\rho<0.85$ falsification threshold. The paper pivots to a methodology
contribution. If pilot correlation falls in $(0.65,0.95)$ we proceed to
the full study as designed. Below $0.65$ the claim is so strongly
supported that the empirical bar can be cleared with reduced scope.

\paragraph{Required submission infrastructure.} Two community artifacts
are required before any leaderboard can report predictive-validity. A
reference shared rule pipeline for judge-independent verification (CAR
computation). A reference adversarial-perturbation suite: 30 base
scenarios with one perturbation per class (paraphrase, identifier
renaming, time-window shifting, distractor injection).

\section{Headline Chart per Study (Source Pointers)}
\label{appx:headline}

For each study, the source report's headline chart that supports the
single most striking finding cited in the main body of this paper.
These chart references are pointers for readers verifying claims; we
do not reproduce the charts here. Each entry cites the source report
where the chart appears.

\begin{itemize}\setlength\itemsep{2pt}
  \item G3 Skills+KP \citep{li2025skillsknowledge}; Token-usage bar chart (Knowledge Plugin vs.\ RAG, 10 scenarios).
  \item G5 Multi-Turn \citep{li2025multiturn}; Per-turn latency line chart by architecture.
  \item G7 Battery \citep{gowda2025agentopsbench}; Optimization ablation table (Naive $\to$ +Parallel $\to$ +Compile $\to$ +Batched $\to$ +Disk-cache).
  \item G8 Transformer \citep{kumar2025transformer}; Pipeline-time vs.\ scenario-count table ($N\in\{10,25,50\}$, baseline vs.\ optimized).
  \item G9 Temporal Cache \citep{merchant2025temporal}; Per-row baseline-vs-optimized latency scatter.
  \item G12 SmartGridBench \citep{bhandari2025smartgridbench}; Core matrix table (seven cells $\times$ judge mean and pass rate).
  \item G14 Conf.~Gate \citep{mazeeva2025skillknowledge}; Threshold sweep table for the gated configuration ($\theta\in\{0.5,\dots,0.95\}$).
  \item G16 FMSR Dispatch \citep{jebbouri2025fmsrbottleneck}; Quantization $\times$ model-family accuracy bar chart vs.\ cloud-hosted Llama-3.3-70B baseline.
  \item G19 TSFM Backend \citep{go2025tsfmprofile}; TTM forecasting stage-breakdown bar chart.
  \item G20 QLoRA \citep{group202025qlora}; 5-dimension $\times$ 4-configuration grouped bar chart of LLM-judge scores.
  \item G21 Profiling \citep{li2025profiling}; Per-scenario latency breakdown table (reasoning-off vs.\ on).
  \item G27 TSFM Optim.\ \citep{vinod2025tsfmoptim}; Ablation incremental-speedup table.
  \item G30 PHMForge \citep{li2025phmforge}; Per-category pass-all-3 table (RAG-fuzzy / RAG-explicit / MCP-fuzzy / MCP-explicit).
  \item G23 Visual Inspect \citep{sheikh2025vi}; 22-scenario sweep table (E2E latency, p50, judge pass) across L0--L3 quantization/serving variants for two VL models.
\end{itemize}

\end{document}